\documentclass[conf]{new-aiaa}
\usepackage[utf8]{inputenc}
\usepackage{fontawesome5}
\usepackage{graphicx}
\usepackage{wrapfig}
\usepackage{float}
\usepackage{tabularx}
\usepackage{booktabs}
\usepackage{amsmath}
\usepackage[version=4]{mhchem}
\usepackage{siunitx}
\usepackage{longtable,tabularx}
\usepackage{url}
\usepackage{caption}
\usepackage{needspace}
\usepackage{multirow}
\usepackage{hyperref}
\hypersetup{
    colorlinks=true,
    urlcolor=blue
}
\usepackage[most]{tcolorbox}     
\usepackage{enumitem}     
\usepackage{placeins}  
\usepackage[table]{xcolor}
\usepackage{afterpage}
\definecolor{promptHead}{HTML}{1F4E79}
\definecolor{promptBody}{HTML}{EAF2F8}
\definecolor{promptCode}{HTML}{F8F1E4}

\definecolor{sampleHead}{HTML}{00695C}
\definecolor{sampleBody}{HTML}{E0F2F1}

\definecolor{qualBody}{HTML}{F8EEBA}
\definecolor{qualHead}{HTML}{8B6F2A}

\definecolor{genHead}{HTML}{B71C1C}
\definecolor{genBody}{HTML}{FFEBEE}
\definecolor{genCode}{HTML}{FBE9E9}

\newtcolorbox{genprompt}[1][]{%
    enhanced, breakable,
    colback=genBody, colframe=genHead,
    colbacktitle=genHead, coltitle=white,
    fonttitle=\bfseries\sffamily, title=#1,
    titlerule=0pt, boxrule=0.6pt, arc=3pt,
    left=10pt, right=10pt, top=8pt, bottom=8pt,
    drop fuzzy shadow=black!25!white,
    pad at break=6pt,
    toprule at break=0.6pt,
    bottomrule at break=0.6pt,
    title after break={\bfseries\sffamily\color{white}#1~\textit{(cont.)}},
}

\definecolor{pilotA}{HTML}{1565C0}        
\definecolor{pilotB}{HTML}{E65100}        
\definecolor{pilotC}{HTML}{1B5E20}        
\definecolor{pilotD}{HTML}{4A148C}        
\definecolor{labelHazard}{HTML}{C62828}
\definecolor{labelWarn}{HTML}{EF6C00}
\definecolor{labelNominal}{HTML}{2E7D32}

\definecolor{chatbg}{RGB}{240,242,245}
\definecolor{inputbg}{RGB}{255,255,255}
\definecolor{outputbg}{RGB}{220,235,255}
\definecolor{headerblue}{RGB}{30,80,160}
\definecolor{warnred}{RGB}{200,50,50}
\definecolor{nomgreen}{RGB}{30,130,60}
\definecolor{metargray}{RGB}{80,80,90}
\definecolor{adsbpurple}{RGB}{100,50,160}
\definecolor{ctaforange}{RGB}{180,90,0}
\definecolor{vlmteal}{RGB}{0,110,120}

\newtcolorbox{vlmprompt}[1][]{%
    enhanced,
    colback=promptBody, colframe=promptHead,
    colbacktitle=promptHead, coltitle=white,
    fonttitle=\bfseries\sffamily, title=#1,
    titlerule=0pt, boxrule=0.6pt, arc=3pt,
    left=10pt, right=10pt, top=8pt, bottom=8pt,
    drop fuzzy shadow=black!25!white,
}

\newtcolorbox{samplebox}[1][]{%
    enhanced,
    colback=sampleBody, colframe=sampleHead,
    colbacktitle=sampleHead, coltitle=white,
    fonttitle=\bfseries\sffamily, title=#1,
    titlerule=0pt, boxrule=0.6pt, arc=3pt,
    left=10pt, right=10pt, top=8pt, bottom=8pt,
    drop fuzzy shadow=black!25!white,
}

\newtcolorbox{qualbox}[1][]{%
    enhanced,
    colback=qualBody, colframe=qualHead,
    colbacktitle=qualHead, coltitle=white,
    fonttitle=\bfseries\sffamily, title=#1,
    titlerule=0pt, boxrule=0.6pt, arc=3pt,
    left=10pt, right=10pt, top=8pt, bottom=8pt,
    drop fuzzy shadow=black!25!white,
}

\newcommand{\spkA}{\textcolor{pilotA}{\faHeadset~\textbf{N910YZ}}}
\newcommand{\spkB}{\textcolor{pilotB}{\faHeadset~\textbf{N602SK}}}

\title{Towards Automated Air Traffic Safety Assessment Around Non-Towered Airports Using Large Language Models}

\author{Torsten Darrell\footnote{Research Intern, Department of Mechanical and Aerospace Engineering, George Washington University, AIAA Student Member.}, Mahyar Ghazanfari\footnote{Ph.D. Student, Department of Mechanical and Aerospace Engineering, George Washington University, AIAA Student Member.}}
\affil{George Washington University, Washington, DC, 20052}

\author{Jordan Kam\footnote{Undergraduate Student, Aerospace Program, University of California, Berkeley, AIAA Student Member.}, Alexandre Bayen\footnote{Full Professor, Department of Electrical Engineering and Computer Science, University of California, Berkeley.}}
\affil{University of California, Berkeley, Berkeley, CA, 94720}

\author{Amin Tabrizian\footnote{Ph.D. Student, Department of Computer Science, George Washington University, AIAA Student Member.} and Peng Wei\footnote{Associate Professor, Department of Mechanical and Aerospace Engineering, George Washington University, AIAA Associate Fellow.}}
\affil{George Washington University, Washington, DC, 20052}

\begin{document}

\maketitle
\textbf{We investigate frameworks for post-flight safety analysis at non-towered airports using large language models (LLMs). Non-towered airports rely on the Common Traffic Advisory Frequency (CTAF) for air traffic coordination and experience frequent near mid-air collisions due to the pilot self-announcement communication protocol.  We propose a general vision-language model (VLM) approach to analyze the transcribed CTAF radio communications in natural language, METeorological Aerodrome Report (METAR) weather data, Automatic Dependent Surveillance-Broadcast (ADS-B) flight trajectories, and Visual Flight Rules sectional charts of the airfield. We provide a preliminary study at Half Moon Bay Airport, with a qualitative real world case study and a quantitative evaluation using a new synthetic dataset of communications and weather modalities. We qualitatively evaluate our framework on real flight data using Gemini~2.5~Pro, demonstrating accurate identification of a right-of-way violation. The synthetic dataset is derived from real examples and includes a 12-category hazard taxonomy, and is used to benchmark three open-source (Qwen~2.5-7B, Mistral-7B, Gemma-2-9B) and three closed-source (GPT-4o, GPT-5.4, Claude Sonnet~4.6) LLM models on the subset of inputs related to CTAF and METAR. Even limited to CTAF and METAR inputs and open source LLMs,  instances of our framework typically achieve a macro $F_1$ score above 0.85 on a binary nominal/danger classification task. Future work includes a quantitative evaluation across all modalities and a larger number of real world examples. Taken together, our results suggest that VLM analysis of safety at non-towered airports may be a valuable future capability.}

\section{Introduction}

Traditional air traffic control (ATC) employs complex decision-making to ensure mid-air collision-avoidance in the United States national airspace system (NAS) \cite{mogford1995complexity}. Airspace operations at non-towered airports follow procedures outlined in the Federal Aviation Administration's (FAA) advisory circular (AC) 90-66C document \cite{FAA_AC90_66C}. This communication is administered between pilots over the common traffic advisory frequency (CTAF) to offer a level of safety in the absence of a human controller. While this collision-avoidance method has been in place for nearly fifty years, its heavy reliance on pilot's voluntary decisions to announce their aircraft's position and intent to others has led to many near mid-air collisions at these non-towered airports \cite{nasa_asrs_nonTower2025}. CTAF's lack of layered safety highlights the shortcomings of purely pilot-to-pilot deconfliction. As the amount of general aviation (GA) in low-altitude airspace increases, the development of automated ATC as a safety layer at these non-towered airports could close this safety critical technology gap.

ATC's deconflition architecture relies on rigid rule-based or data-driven systems that function well only when inputs are complete and precisely formatted (e.g., sensor data or communication) \cite{NAS_InTimeAviationSafety2022}. Furthermore, these architectures are primarily human-in-the-loop. However, non-towered airport operations are inherently unstructured. CTAF exchanges are spontaneous, abbreviated, and often incomplete radio calls whose meaning depends heavily on contextual cues such as traffic position, weather, and visual conditions \cite{FAA_AC90_66C}. Non-towered airports also typically lack the infrastructure required to log flight data essential for monitoring airspace operations.  

Recent advances in artificial intelligence (AI) have produced remarkable progress across audio, vision, and language domains, enabling large language models (LLMs) and vision-language models (VLMs) to perform a broad range of translation, comprehension, and reasoning tasks \cite{vaswani2017attention, cui2025recent, yin2024survey, stahlberg2020neural, plaat2024reasoning}, opening the possibility they could handle the unstructured, heterogenous, and semantically-rich nature of non-towered ATC data. LLMs have demonstrated strong reasoning capabilities over natural language, structured data, and contextual information \cite{cheng2024callmenn}. Pre-trained on large-scale text corpora, these models learn relationships among language, numerical data, and higher-level concepts, enabling them to infer deeper contextual information such as intent. VLMs extend these capabilities to multimodal settings by jointly reasoning over visual and textual inputs, making them well suited for aviation applications involving heterogeneous data sources such as flight trajectories, airport diagrams, weather information, and pilot communication transcripts. 

We propose LLM-based models for  post-flight safety assessments of airspace operations surrounding non-towered airports. We consider a general VLM approach to analyze the relevant aviation signals including transcribed CTAF radio communications in natural language, METeorological Aerodrome Report (METAR) weather data, Automatic Dependent Surveillance-Broadcast (ADS-B) flight trajectories, and Visual Flight Rules (VFR) sectional charts of the airfield.
 A qualitative experiment using Half Moon Bay Airport (KHAF) as a case study suggests that a contemporary VLM can discriminate safe vs unsafe conditions when prompted with these inputs.  Initial quantitative evaluations of model performance when prompted soley with CTAF and METAR data confirm performance across a variety of models and parameter setttings. Future work includes a quantitative evaluation across all modalities and a larger number of real world examples.
We believe that LLM and VLM based models can serve as an important element to support safe airspace operations around non-towered airports and/or in general scenarios where human air traffic controllers are unavailable.

\section{Related Work}

While there is a long history of automated collision avoidance, including terrain avoidance \cite{kuchar2007traffic} and head-to-head collision resolution systems \cite{al2010probability} that have been deployed in modern aviation based on explicit predictive models and control theory \cite{kochenderfer2012next}, analysis of natural language CTAF communications for aircraft safety or intent prediction has been relatively underexplored. Prior work has considered natural language processing for pilot communications, including extensive investigation of Automatic Speech Recognition (ASR) for automatic transcription, and the development of aviation safety databases and incident evaluations \cite{paradis2025kaona}. Other work using LLMs, such as ChatGPT, have been used for summarization of historical ground delay program (GDP) data to support optimal air traffic flow management in strategic decision making \cite{abdulhak2024chatatc}. Lastly, prior research has studied non-towered airports and modeled ATC as a Markov Decision Process (MDP) towards autonomous ATC of GA operations \cite{mahboubi2015autonomous}.

Prior works have used LLMs for safety report summarization of aviation report data. LLMs have been used to summarize incident narratives from the Aviation Safety Reporting System (ASRS) \cite{tikayat2023examining} and extract causal patterns related to human factors, while \cite{nielsen2024towards} investigated domain adaptation of transformer-based architectures for documentation such as Letters of Agreements. Both work demonstrating improved accuracy in domain-specific classification. The AviationGPT framework \cite{wang2024aviationgpt} further introduced a domain-tuned LLaMA-based \cite{touvron2023llama} LLM that achieved substantial performance gains on aviation text corpora. In parallel, \cite{gao2024workloadgpt} employed an LLM to detect pilot workload levels from multimodal physiological and behavioral data. Beyond these textual and design applications, \cite{tabrizian2025chain} proposed a chain-of-thought-based LLM flight planner for eVTOL routing under wind hazards, achieving high plan validity while incorporating human preferences via natural-language input. \cite{andriuvskevivcius2024automatic} on the other hand, introduced an embodied ATC agent using LLMs with function-calling and an experience library to autonomously resolve multi-aircraft conflicts in simulated airspace \cite{su2025flight}, achieving near-human reasoning performance. 

More recently, \cite{abdulhak2024chatatc} developed CHATATC, a fine-tuned conversational agent trained on over 80,000 GDP issuances to support strategic air-traffic-flow management. The study demonstrated that LLM-based summarization and dialogue systems can assist Traffic Managers by retrieving, synthesizing, and contextualizing historical NAS data for decision support. In addition, \cite{sangeetha2025language} introduced a multimodal model that integrates ASR and LLM-based intent extraction with spatial reasoning for goal prediction in untowered airspace, showing that language-conditioned multimodal fusion significantly reduces goal-prediction error. Collectively, these studies highlight the expanding role of LLMs as reasoning and decision-support systems across aviation, covering domains such as safety analysis, workload estimation, multimodal intent prediction, route generation, and traffic-flow management. Yet, existing efforts primarily focus on prediction and decision-support tasks rather than 
understanding 
pilot communications.

\section{Framework}

\begin{figure}[t]
    \centering
    \fbox{%
        \includegraphics[width=1\linewidth]{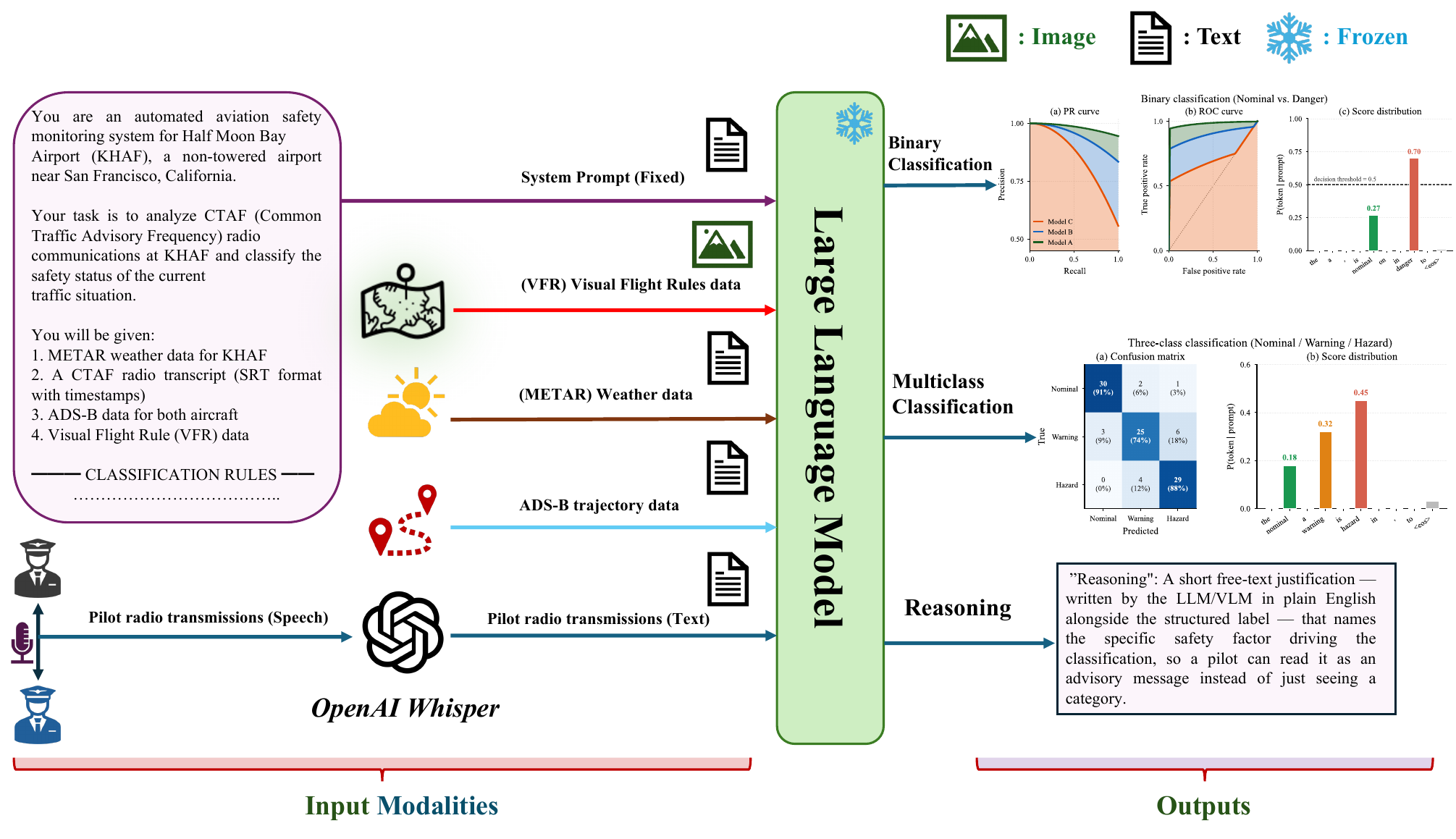}%
    }
    \caption{Framework overview. Pilot radio communications are
    transcribed by Whisper~Large-v3 and passed to a frozen LLM alongside METAR weather, ADS-B trajectory, VFR sectional chart of the non-towered airfield, and system prompt text. The model emits both
    a structured safety classification (binary, with a three-class
    breakdown in Appendix~\ref{app:3class}) and a free-text reasoning
    string that can be surfaced to the pilot as a CTAF-style advisory.}
    \label{fig:1}
\end{figure}

\subsection{Overview}
\label{sec:method:overview}

Our framework (Fig.~\ref{fig:1}) consists of a LLM prompted with a heterogeneous set of data modalities based on aviation inputs which generates both a structured text safety classification and a free-text reasoning output. Five inputs are concatenated into a single prompt: a fixed task prompt (text), METAR weather data (text), ADS-B flight trajectories (text), transcribed CTAF communications between pilots (text), and a VFR sectional chart of the airfield (image). The model's structured output is a binary nominal-versus-danger classifier; we also evaluate using a finer-grained three-class safety classifier, reported in Appendix~\ref{app:3class}. 

The LLM is frozen throughout this work, no weights are updated, and no fine-tuning is performed using the synthetic or real flight data.  The model's behavior is shaped entirely through prompt engineering, in-context learning, and the choice of reasoning. This design choice preserves modularity, and the LLM used can be swapped or upgraded independently of the prompt template or the ASR component. We evaluate using a variety of LLM/VLM models as described below. The model implements a mapping:
\[
f_{\theta} : (x_{\text{prompt}}, x_{\text{CTAF}}, x_{\text{METAR}},
              x_{\text{ADS-B}}, x_{\text{VFR}}) \rightarrow (y, \alpha)
\]
where $y$ is the predicted safety category and $\alpha$ is the
free-text reasoning.

Whisper performs a deterministic mapping $g_{\phi}: x_{\text{audio}} \rightarrow x_{\text{CTAF}}$, transcribing raw radio audio (with timestamps preserved in SubRip Text format) into the textual representation later provided to the model.

\begin{figure*}[!h]
\begin{qualbox}
\textbf{Input Modalities}
 
\smallskip
\noindent
\begin{tcolorbox}[
    enhanced,
    colback=sampleBody, colframe=sampleHead,
    boxrule=0.5pt, arc=2pt,
    left=4pt, right=4pt, top=4pt, bottom=4pt]
  {\footnotesize\bfseries\color{sampleHead}\faMap\;VFR Sectional Chart --- KHAF (Half Moon Bay Airport)}\\[4pt]
  \includegraphics[width=\linewidth, trim=0 60 0 60, clip]%
    {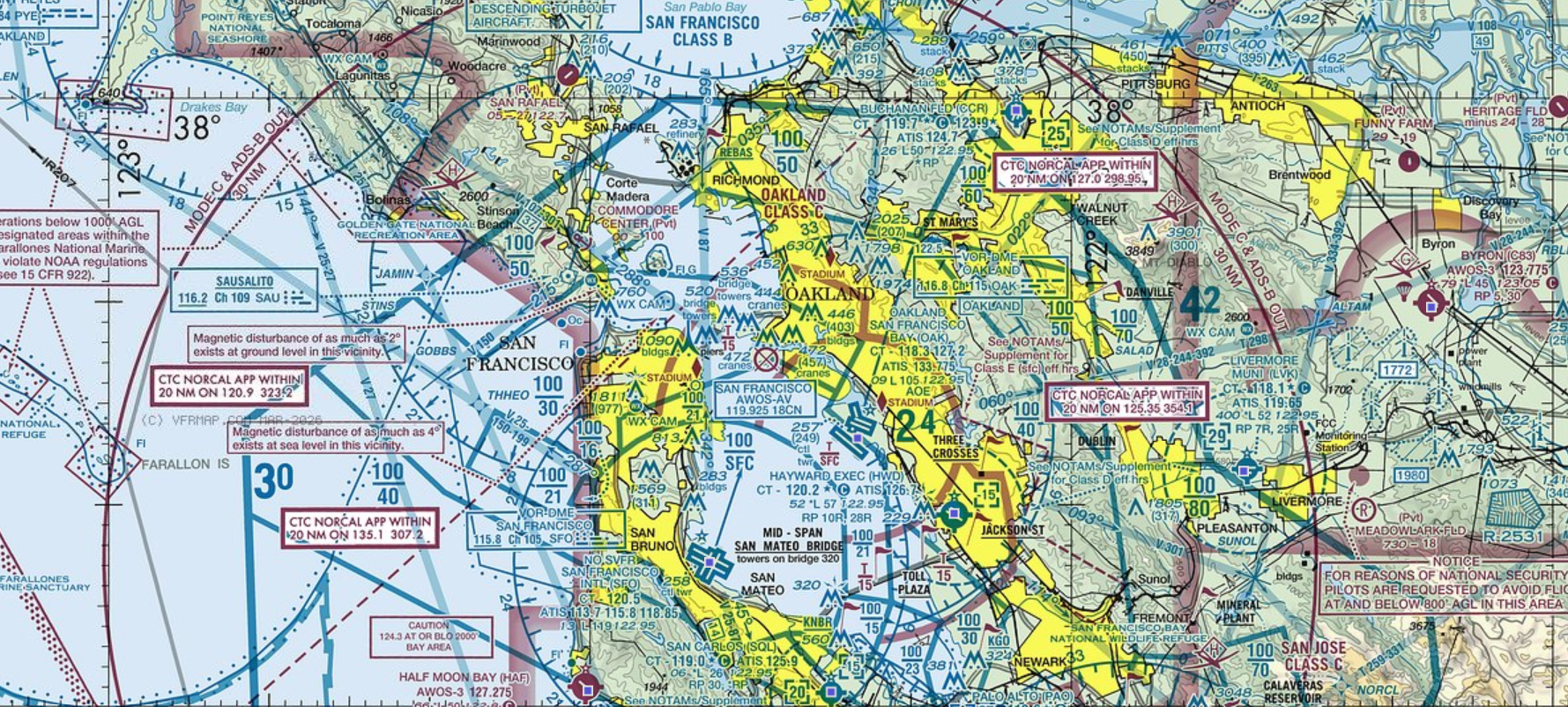}
\end{tcolorbox}
 
\smallskip
\noindent
\begin{tabular}{@{}p{0.48\linewidth}@{\hspace{0.02\linewidth}}p{0.48\linewidth}@{}}
\begin{tcolorbox}[
    enhanced,
    colback=promptBody, colframe=promptHead,
    boxrule=0.5pt, arc=2pt,
    left=5pt, right=5pt, top=4pt, bottom=4pt]
  {\footnotesize\bfseries\color{promptHead}\faCloudSun\;METAR --- KHAF}\\[4pt]
  {\ttfamily\small KHAF 041655Z AUTO 27005KT 9SM BKN060 14/13 A2999 RMK AO2}\\[4pt]
  {\small Wind 270°/5~kt $\cdot$ Visibility 9~SM $\cdot$
   BKN ceiling 6{,}000~ft $\cdot$
   Temp 14°C / Dewpoint 13°C $\cdot$
   Altimeter 29.99~inHg}
\end{tcolorbox}
&
\begin{tcolorbox}[
    enhanced,
    colback=promptBody, colframe=promptHead,
    boxrule=0.5pt, arc=2pt,
    left=5pt, right=5pt, top=4pt, bottom=4pt]
  {\footnotesize\bfseries\color{promptHead}\faSatelliteDish\;ADS-B State Data}\\[4pt]
  {\small
  \begin{tabular}{@{}lll@{}}
    \toprule
    & \textbf{N100AB} & \textbf{N200CD}\\
    \midrule
    Altitude (ft)     & 1{,}800$\downarrow$ & 900$\uparrow$\\
    Speed (kt)        & 75   & 60\\
    Heading (°)       & 301  & 210\\
    Vert.\ rate (fpm) & $-$600 & $+$200\\
    \bottomrule
  \end{tabular}}\\[4pt]
  {\small \textcolor{pilotC}{\faHeadset~\textbf{N100AB}}: straight-in RNAV\quad
          \textcolor{pilotD}{\faHeadset~\textbf{N200CD}}: right-base turn}
\end{tcolorbox}
\end{tabular}
 
\medskip
 
\textbf{CTAF Transcript} (Whisper Large-v3 ASR $\cdot$ KHAF 122.8~MHz)\\[2pt]
{\footnotesize\itshape
Standard call frame ``Half Moon Bay traffic, \ldots, Half Moon Bay''
omitted for brevity.\quad
\textcolor{pilotC}{\faHeadset~\textbf{N100AB}} = straight-in RNAV\quad
\textcolor{pilotD}{\faHeadset~\textbf{N200CD}} = VFR pattern\quad
\textcolor{gray}{$\bullet$~italics} = other traffic}
 
\smallskip
{\small
\begin{tabular}{@{}r@{\hspace{0.6em}}p{0.87\linewidth}@{}}
\texttt{01:13} & \textcolor{pilotD}{\faHeadset~\textbf{N200CD}} ``turning right crosswind 30.''\\
\texttt{02:52} & \textcolor{pilotD}{\faHeadset~\textbf{N200CD}} ``turning right base, 30.''\\
\texttt{03:44} & \textcolor{pilotC}{\faHeadset~\textbf{N100AB}} ``12 miles to the southeast, 3{,}400~ft, descending on the instrument approach for runway 30, full stop.''\\
\texttt{04:10} & \textcolor{pilotC}{\faHeadset~\textbf{N100AB}} ``coming up on 6-mile final for 30, 1{,}500~ft.''\\
\texttt{04:43} & \textcolor{pilotC}{\faHeadset~\textbf{N100AB}} ``10 miles to the southeast, 3{,}300~ft, straight in on the instrument approach runway 30, full stop.''\\
\texttt{05:49} & \textcolor{pilotC}{\faHeadset~\textbf{N100AB}} ``4-mile final.''\\
\texttt{05:55} & \textcolor{pilotC}{\faHeadset~\textbf{N100AB}} ``7 miles to the southeast, 2{,}600~ft, straight in on the instrument for runway 30, full stop.''\\
\texttt{06:11} & \textcolor{pilotD}{\faHeadset~\textbf{N200CD}} ``turning crosswind.''\\
\texttt{06:46} & \textcolor{pilotC}{\faHeadset~\textbf{N100AB}} ``5 miles southeast, 1{,}900~ft, straight in instrument approach, runway 30, full stop.''\\
\texttt{07:40} & \textcolor{pilotD}{\faHeadset~\textbf{N200CD}} ``on downwind --- will turn base after the traffic on short final.''\\
\texttt{07:49} & \textcolor{pilotC}{\faHeadset~\textbf{N100AB}} ``3 miles to the southeast, straight in on the instrument approach, full stop, runway 30.''\\
\texttt{08:20} & \textcolor{pilotD}{\faHeadset~\textbf{N200CD}} ``turning right base, 30.''\\
\texttt{08:31} & \textcolor{pilotC}{\faHeadset~\textbf{N100AB}} \textbf{\textcolor{labelHazard}{``2-mile final for runway 30 --- traffic turning final in front of us; we're gonna have to go around if you're not off that runway.''}}\\
\texttt{08:49} & \textcolor{pilotC}{\faHeadset~\textbf{N100AB}} ``we see you deviating off to the right now, 1$\frac{1}{2}$-mile final, full stop for 30.''\\
\texttt{10:29} & \textcolor{pilotD}{\faHeadset~\textbf{N200CD}} ``turning final, about a 2-mile final.''\\
\end{tabular}}
 
\end{qualbox}
\caption{Qualitative results of the VLM safety analysis pipeline at KHAF, using a VFR sectional, METAR, ADS-B state data, and a Whisper-transcribed CTAF feed. Gemini identifies a right-of-way violation on final approach, resolved by a go-around warning. \textit{Call signs shown are random placeholders, not the real call signs; all other data is unmodified.}}
\label{fig:qualitative_khaf}
\end{figure*}

\begin{figure*}[t]\ContinuedFloat
\begin{qualbox}
\textbf{CTAF Transcript} \textit{(continued)}
 
\smallskip
{\small
\begin{tabular}{@{}r@{\hspace{0.6em}}p{0.87\linewidth}@{}}
\texttt{12:49} & \textcolor{gray}{\textit{Cherokee 1AA}} ``4 miles to the southeast at 2{,}000~ft descending, will fly over the ocean, cross midfield into the right downwind.''\\
\texttt{13:21} & \textcolor{gray}{\textit{Skyhawk 2BB}} ``10 miles to the southeast on RNAV 30 inbound.''\\
\texttt{14:47} & \textcolor{pilotD}{\faHeadset~\textbf{N200CD}} ``turning right downwind, right traffic 30.''\\
\texttt{15:21} & \textcolor{gray}{\textit{Skyhawk 2BB}} ``5 miles to the southeast on RNAV 30, full stop.''\\
\texttt{15:50} & \textcolor{gray}{\textit{Archer 3CC}} ``6 miles to the southeast, maneuvering over water, then crossing midfield to enter right downwind 30.''\\
\texttt{16:09} & \textcolor{gray}{\textit{Cherokee 1AA}} ``overhead midfield entering right downwind 30 --- continuing climb, departing area to the south, 1{,}500~ft.''\\
\texttt{16:23} & \textcolor{pilotD}{\faHeadset~\textbf{N200CD}} ``right to base, 30.''\\
\texttt{16:43} & \textcolor{gray}{\textit{Skyhawk 2BB}} ``on the RNAV, is that you at 2{,}300?'' \textcolor{gray}{\textit{Skyhawk 2BB:}} ``Affirm, on the RNAV 30.''\\
\texttt{17:00} & \textcolor{gray}{\textit{Skyhawk 2BB}} ``we'll be going under you.''\\
\texttt{17:35} & \textcolor{gray}{\textit{Skyhawk 2BB}} ``5 miles to the southeast on RNAV 30, full stop.''\\
\texttt{18:45} & \textcolor{gray}{\textit{Skyhawk 2BB}} ``3 miles final on 30.''\\
\texttt{19:10} & \textcolor{gray}{\textit{Archer 3CC}} ``offshore, turning to cross midfield and enter the right downwind 30.''\\
\texttt{19:27} & \textcolor{gray}{\textit{Skyhawk 2BB}} ``on the RNAV 30.''\\
\texttt{20:10} & \textcolor{gray}{\textit{Archer 3CC}} ``crossing midfield to enter right downwind 30.''\\
\texttt{20:23} & \textcolor{gray}{\textit{Skyhawk 2BB}} ``short final 30.''\\
\texttt{20:43} & \textcolor{gray}{\textit{Archer 3CC}} ``turning onto the downwind 30.''\\
\texttt{20:54} & \textcolor{pilotD}{\faHeadset~\textbf{N200CD}} ``turning downwind --- I have the traffic on the downwind.''\\
\texttt{21:39} & \textcolor{gray}{\textit{Archer 3CC}} ``turning right base 30.''\\
\texttt{22:28} & \textcolor{gray}{\textit{Archer 3CC}} ``on final 30.''\\
\texttt{23:01} & \textcolor{pilotD}{\faHeadset~\textbf{N200CD}} ``turning right base 30.''\\
\texttt{23:17} & \textcolor{gray}{\textit{Unknown}} ``2 miles to the north, transitioning the area at 2{,}500~ft.''\\
\texttt{23:44} & \textcolor{pilotD}{\faHeadset~\textbf{N200CD}} ``turning final 30.''\\
\texttt{26:55} & \textcolor{pilotD}{\faHeadset~\textbf{N200CD}} ``turning crosswind to right traffic 30.''\\
\texttt{28:32} & \textcolor{pilotD}{\faHeadset~\textbf{N200CD}} ``turning right base 30.''\\
\texttt{29:29} & \textcolor{gray}{\textit{Mania 4DD}} ``right downwind 30, departing the pattern to the southeast.''\\
\texttt{30:58} & \textcolor{gray}{\textit{Desmond 5EE}} ``about 2 miles to the airfield, making right traffic, runway 30.''\\
\end{tabular}}
 
\medskip
 

\textbf{VLM Safety Analysis}
  
\medskip
\begin{tcolorbox}[
    enhanced,
    colback=labelHazard!8, colframe=labelHazard!60!black,
    boxrule=0.5pt, arc=2pt,
    left=10pt, right=10pt, top=6pt, bottom=6pt]
\textbf{Primary Conflict (07:40--08:51) --- Right-of-Way Violation on Final Approach}
 
\smallskip
At \texttt{07:40}, \textcolor{pilotD}{\textbf{N200CD}} reports downwind
with an ambiguous sequencing call (``will turn base after the traffic
on short final''), suggesting awareness of the inbound but unclear
intent. At \texttt{08:20}, N200CD turns base, misjudging the closure
rate of the established straight-in \textcolor{pilotC}{\textbf{N100AB}}.
At \texttt{08:31}, N100AB on a 2-mile final issues a go-around threat,
directly resolving the conflict: N200CD immediately deviates right and
N100AB confirms visual separation and continues to a full stop.
 
\medskip
\textit{Instructional point.} An aircraft established on final approach
has right-of-way over aircraft in the traffic pattern (FAA\,AC\,90-66C).
N200CD failed to yield; N100AB demonstrated correct airmanship by clearly
announcing the conflict, which prompted immediate corrective action and
prevented a potential near mid-air collision.
 
\medskip
\begin{tcolorbox}[
    enhanced,
    colback=labelHazard!15, colframe=labelHazard,
    boxrule=0.5pt, arc=2pt,
    left=10pt, right=10pt, top=4pt, bottom=4pt]
\textbf{Label:}~\textcolor{labelHazard}{\textsc{Hazard}}\quad\textbar\quad
\textbf{Confidence:}~0.91\quad\textbar\quad
\textbf{Hazard type:}~\textit{simultaneous\_final / right-of-way violation}\quad\textbar\quad
\textbf{Model:}~\texttt{Gemini-2.5-Pro}
\end{tcolorbox}
 
\end{tcolorbox}
 
\end{qualbox}
\end{figure*}
\FloatBarrier

\subsection{Qualitative Example}
\label{sec:qualitative}

We probed the framework's ability to perform safety assessments using an example at KHAF using four aviation data sources across two modalities (text and image) from real flight data. The data sources include ADS-B trajectory data, CTAF communications, METAR weather data, and a VFR sectional chart. Figure~\ref{fig:qualitative_khaf} presents a live CTAF recording from KHAF transcribed by Whisper~Large-v3 and analyzed by Gemini~2.5~Pro. By inputting KHAF's VFR sectional chart as an image to the model, this qualitative study serves as a preliminary investigation into the effectiveness of using VLMs for airspace safety analysis. The same figure also demonstrates the complete process of testing the qualitative real-world scenario at KHAF. Gemini~2.5~Pro was capable of accurately identifying a right-of-way violation that led to a near mid-air collision. 

\section{Synthetic Dataset and METAR+CTAF Benchmark Scenarios}
\label{sec:method:dataset}

\subsection{Dataset}
For our quantitative study, we restrict the inputs to METAR weather data and CTAF communication transcripts, which are processed using a LLM. 
We develop and evaluate on \texttt{CTAF-KHAF-Synthetic}, a synthetic benchmark of 100 flight-operations scenarios at KHAF non-towered airport derived from real data examples. The full generator prompt is reproduced in Appendix~\ref{app:gen_prompt}. The 12 hazard categories are drawn from incident patterns documented in FAA AC~90-66C~\cite{FAA_AC90_66C} and the NASA ASRS non-tower report set~\cite{nasa_asrs_nonTower2025}, so each scenario type corresponds to a class of unsafe (or nominal) situation that actually occurs at non-towered airports. Pilot phraseology follows AC~90-66C
self-announcement conventions, and METAR and per-aircraft ADS-B
values use the same fields, units, and ranges observed in real KHAF
operations.

Each scenario is first assigned a binary ground-truth safety label of either \texttt{nominal} or \texttt{danger}, reflecting whether the traffic situation contains any operational safety concern. Each scenario additionally includes a hazard-type label drawn from a 12-category taxonomy spanning communication gaps, pattern conflicts, runway incursions, instrument flight rules or VFR misalignment, and nominal operations, along with raw and decoded METAR text, a CTAF radio transcript in SRT format, and multi-voice audio synthesized from the transcript using OpenAI's TTS-1-HD. For the finer-grained three-class formulation used in Appendix~\ref{app:3class}, the \texttt{danger} class is further subdivided into \texttt{warning} and \texttt{hazard} based on collision imminence. A \texttt{warning} corresponds to a potentially unsafe situation that pilots can still resolve through standard advisory actions, whereas a \texttt{hazard} represents an imminent collision risk or serious operational conflict. Under this formulation, the labels are nearly balanced (33~nominal, 34~warning, 33~hazard). The scenarios were constructed from the 12-category hazard taxonomy by composing CTAF radio call sequences that exhibit each target safety condition and a representative METAR. Each transcript was then synthesized into multi-voice audio using OpenAI's TTS-1-HD, with a distinct voice assigned to each aircraft on the frequency so that speaker attribution is preserved in the audio.
The complete dataset and code are available \href{https://github.com/Mahyar-GH79/Automated-Multimodal-Analysis-of-Air-Traffic-around-Non-Towered-Airports-using-Large-Language-Models}{here}.

\begin{figure}[b!]
\centering
\begin{minipage}{0.95\linewidth}

\begin{samplebox}[Sample Scenario --- S003: Simultaneous Final on Runway 30]

\textbf{Setting.} Half Moon Bay Airport (KHAF), runway 30, right-traffic pattern.

\medskip
\textbf{METAR.} \texttt{KHAF 142135Z AUTO 18005KT 5SM -BR FEW010 BKN020 18/16 A2999 RMK AO2}\\
\textit{Decoded:} Marginal VFR --- 5~SM visibility in mist, broken ceiling
at 2{,}000~ft, wind 180$^{\circ}$ at 5~kt, 18$^{\circ}$C / dewpoint
16$^{\circ}$C.

\medskip
\textbf{Aircraft on frequency.}
\begin{itemize}[leftmargin=1.4em, topsep=2pt, itemsep=2pt]
  \item \textcolor{pilotA}{$\blacktriangleright$~\textbf{N910YZ}} --- Cessna~172, two-mile
        straight-in RNAV final runway 30, \emph{full stop}.\\
        \textit{ADS-B (t=0):} 37.4967$^{\circ}$N, 122.4644$^{\circ}$W;
        800~ft~MSL; heading~300$^{\circ}$; 85~kt.
  \item \textcolor{pilotB}{$\blacklozenge$~\textbf{N602SK}} --- Piper~Seneca, right base
        runway 30, \emph{touch-and-go}.\\
        \textit{ADS-B (t=0):} 37.5147$^{\circ}$N, 122.4756$^{\circ}$W;
        900~ft~MSL; heading~210$^{\circ}$; 85~kt.
\end{itemize}

\medskip
\textbf{CTAF transcript.}
{\footnotesize\itshape The standard CTAF call frame
``Half Moon Bay traffic, \ldots, Half Moon Bay'' is omitted for
brevity.}

\smallskip
{\small
\begin{tabular}{@{}r@{\hspace{0.7em}}p{0.80\linewidth}@{}}
\texttt{00:00.0} & \spkB~``right base runway three zero, touch and go.''\\
\texttt{00:04.8} & \spkA~``two-mile straight-in RNAV final runway three zero, full stop.''\\
\texttt{00:10.1} & \spkB~``turning right final runway three zero, touch and go.''\\
\texttt{00:15.3} & \spkA~``one-and-a-half-mile final runway three zero, full stop.''\\
\texttt{00:21.0} & \spkB~``short final runway three zero, touch and go.''\\
\texttt{00:25.5} & \spkA~``one-mile final runway three zero, full stop.''\\
\texttt{00:31.0} & \spkA~``traffic on short final, say position.''\\
\texttt{00:36.1} & \spkB~``on short final three zero, negative contact.''\\
\texttt{00:41.0} & \spkA~``half-mile final, I have traffic now, you're directly below me.''\\
\texttt{00:46.2} & \spkB~``traffic in sight now, you're overtaking us on final.''\\
\texttt{00:51.3} & \spkA~``\textbf{going around} runway three zero, traffic conflict on final.''\\
\texttt{00:57.1} & \spkB~``continuing runway three zero, \textbf{near midair} on short final.''\\
\end{tabular}}

\textbf{Ground-truth advisory} (FAA-conforming).
\begin{quote}\small\itshape
``Traffic alert: N910YZ, Cessna one-mile straight-in final Runway Three~Zero,
and N602SK, Seneca short final Runway Three~Zero, converging on the same
runway. N910YZ, go around immediately; N602SK, continue landing or clear
the runway without delay; both aircraft maintain visual separation.''
\end{quote}

\medskip
\begin{tcolorbox}[
    colback=labelHazard!10, colframe=labelHazard!70!black,
    boxrule=0.5pt, arc=2pt,
    left=10pt, right=10pt, top=4pt, bottom=4pt]
\textbf{Ground-truth label:}~\textcolor{labelHazard}{\textsc{Hazard}}\quad\textbar\quad
\textbf{Hazard type:}~\textit{simultaneous\_final}\quad\textbar\quad
\textbf{Source:} CTAF-KHAF benchmark, scenario \texttt{S003}
\end{tcolorbox}
\end{samplebox}

\end{minipage}
\end{figure}

\begin{figure}[b!]
\centering
\begin{minipage}{0.95\linewidth}

\begin{vlmprompt}[System Prompt --- Binary Safety Classification]
You are an automated aviation safety monitoring system for Half Moon Bay
Airport (KHAF), a non-towered airport near San Francisco, California. Your
task is to analyze CTAF (Common Traffic Advisory Frequency) radio
communications at KHAF and classify the safety status of the current
traffic situation.

\medskip
\textbf{Inputs}
\begin{itemize}[leftmargin=1.4em, topsep=2pt, itemsep=1pt]
    \item METAR weather data for KHAF (raw + decoded)
    \item CTAF radio transcript (SRT format with timestamps)
\end{itemize}

\medskip
\textbf{Task.} Classify the situation as exactly one of \texttt{nominal}
or \texttt{danger}.

\medskip
\textbf{NOMINAL ---} all is well.
\begin{itemize}[leftmargin=1.4em, topsep=2pt, itemsep=1pt]
    \item All required position calls are present
          (crosswind, downwind, base, final)
    \item Traffic is sequenced and separated with no conflicts
    \item Weather is VMC and appropriate for operations
    \item Single aircraft announcing each leg, no other traffic
\end{itemize}

\medskip
\textbf{DANGER ---} any potential or imminent safety issue. Use
\texttt{danger} whenever there is \emph{any} conflict, communication
gap, or unsafe condition:
\begin{itemize}[leftmargin=1.4em, topsep=2pt, itemsep=1pt]
    \item Communication gaps: missing position calls, NORDO traffic,
          delayed announcements
    \item Pattern conflicts: converging traffic, wrong-runway calls,
          improper entries
    \item Active conflicts: simultaneous final, runway incursions,
          mid-air risk
    \item Weather mismatches: VFR pilot inadvertently in IMC
    \item Late or omitted go-around announcements
    \item Any situation a CTAF advisory would flag as caution, alert,
          or emergency
\end{itemize}
\textit{Key question.} ``Would a CTAF advisory flag this for any reason
(caution, alert, or emergency)?'' If yes $\Rightarrow$ \textbf{danger}.

\medskip
\textbf{CTAF rules} (FAA AC~90-66C~\cite{FAA_AC90_66C}).
\begin{itemize}[leftmargin=1.4em, topsep=2pt, itemsep=1pt]
    \item Pilots must self-announce: crosswind, downwind, base, final,
          runway clear
    \item Straight-in: announce at 10, 5, and 3~NM
    \item Go-around must be announced immediately
    \item No ATC --- pilots are solely responsible for separation
\end{itemize}

\medskip
\textbf{Output format.} Respond with \emph{only} the following JSON, no
other text:

\begin{tcolorbox}[
    colback=promptCode, colframe=promptHead!60!black,
    boxrule=0.3pt, arc=2pt,
    left=10pt, right=10pt, top=4pt, bottom=4pt,
    fontupper=\ttfamily\small]
\{\\
\quad "label": "<nominal | danger>",\\
\quad "confidence": <0.0--1.0>,\\
\quad "reasoning": "<one sentence stating the key safety factor>"\\
\}
\end{tcolorbox}
\end{vlmprompt}

\end{minipage}
\end{figure}

Every scenario including its ground-truth safety
label, hazard-type label, and FAA-conforming advisory text was
reviewed by human experts with GA pilot flight experience to
verify that the assigned label is unambiguous and that the radio
call sequence would be plausible in real CTAF operations. Six held-out scenarios, two per class, are reserved as in-context learning (ICL) examples and are excluded from evaluation. The remaining 94 scenarios constitute the test set used in every experiment reported in this article. A representative sample is shown in the S003: Simultaneous Final on Runway 30 box below, illustrating the multi-aircraft conversational structure typical of CTAF traffic, the natural co-occurrence of incomplete or conflicting position calls, and the FAA-conforming advisory text that defines the ground-truth label. METAR codes provide weather
information to pilots operating near an airfield, assisting with
flight planning and instrument adjustments. The corresponding synthetic METAR text data was sourced from National Oceanographic and Atmospheric Association's Aviation Weather Center~\cite{aviationweather}.

\subsection{LLM Evaluation}
\label{sec:method:vlm}

We benchmark six LLMs spanning open-source and closed-source models quantitatively using the synthetic dataset.
The three open-source LLMs include: Qwen~2.5-7B-Instruct~\cite{team2024qwen2}, Mistral-7B-Instruct-v0.3~\cite{jiang2023mistral}, and Gemma-2-9B-IT~\cite{team2024gemma2}. The open-source models are run locally in fp16 with no quantization. The three closed-source models: GPT-4o, GPT-5.4, and Claude Sonnet~4.6 are accessed through their respective HTTP APIs. All six LLMs are frozen at evaluation time. The qualitative experiment is run using the Gemini~2.5~Pro model~\cite{comanici2025gemini}. Each scenario is evaluated under three prompting strategies that vary the number of examples. Zero-shot: The system prompt and target scenario only, no exemplars. One-shot: One held-out exemplar per class (3~total). Few-shot: Two held-out exemplars per class (6~total). Exemplars are drawn from the fixed 6-scenario ICL pool and never appear
in the test set. The same prompt template is run under two reasoning protocols. In direct prompting the model emits the structured JSON in a
single turn. In chain-of-thought (CoT) prompting~\cite{wei2022chain}, a first turn elicits step-by-step
reasoning (``Before classifying, reason step by step\ldots'') and a second turn extracts the JSON from that reasoning.

Each combination of model, strategy, and reasoning protocol constitutes one evaluation condition, yielding a $6\times3\times2 = 36$-condition design per task framing. A fixed system prompt anchors the LLM in the safety-classification
task. It establishes the airport, runway, and traffic-pattern
conventions following FAA AC~90-66C~\cite{FAA_AC90_66C}, defined each safety class with a list of distinguishing criteria. The prompt then prescribes a strict JSON output schema. The complete prompt for the binary task is
shown in the \texttt{System Prompt} box below. A three-class prompt used in Appendix~\ref{app:3class} follows the same structure but splits the \texttt{danger} class into \texttt{warning} and \texttt{hazard}. Every prediction is parsed from a JSON object with three fields: \texttt{label} (the categorical safety class), \texttt{confidence} (a self-reported scalar in $[0,1]$), and \texttt{reasoning} (a one-sentence free-text advisory in FAA-conforming phraseology). The structured fields drive quantitative
evaluation; the \texttt{reasoning} string is a human-readable advisory
that can be displayed to pilots directly.


\section{Quantitative Results}
\label{sec:exp}

\begin{figure}[b!]
    \centering
    \includegraphics[width=\linewidth]{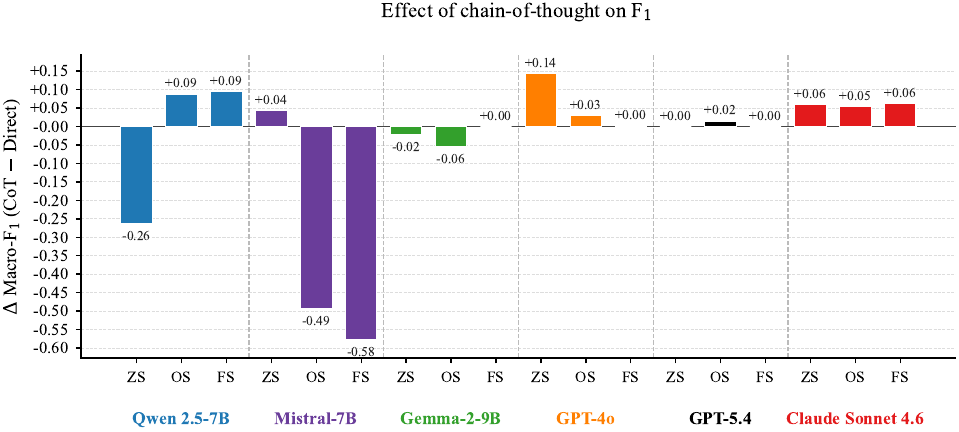}
    \caption{Per-condition macro-$F_1$ change from adding chain-of-thought
    on the binary task. Mistral's collapse is now visible at every
    non-zero-shot setting; GPT-4o and Claude show modest gains.}
    \label{fig:cot_binary}
\end{figure}

We evaluate all six models on the 94-scenario test split of \texttt{CTAF-KHAF-Synthetic} across 36 evaluation conditions (six models $\times$ three prompting strategies $\times$ two reasoning protocols). The primary task is binary safety classification (\texttt{nominal}/\texttt{danger}), while a finer-grained three-class formulation (\texttt{nominal}/\texttt{warning}/\texttt{hazard}) is reported in Appendix~\ref{app:3class}. Additional robustness ablations targeting ASR quality, audio noise, and transcript masking are reported in Appendix~\ref{app:ablations}. Table~\ref{tab:main_binary} reports macro-$F_1$, accuracy, and Area Under the Receiver Operating Characteristic Curve (AUROC) across all evaluation conditions, while Table~\ref{tab:perclass_binary} reports per-class $F_1$ scores. Every LLM exceeds macro-$F_1 = 0.85$ in its best configuration; the strongest configuration is Qwen-2.5-7B at zero-shot direct ($F_1 = 0.964$, $\text{AUROC} = 0.995$), with Claude Sonnet~4.6 close behind at few-shot~+~CoT ($F_1 = 0.952$). GPT-4o and Gemma-2-9B both reach $F_1 \approx 0.93$. Most models approach near-ceiling performance on the binary task, although notable failures remain for Mistral-7B under one-shot and few-shot CoT prompting, where macro-$F_1$ drops to approximately $0.30$.

\subsection{Effect of ICL and CoT}
\begin{figure}[h!]
    \centering
    \includegraphics[width=\linewidth]{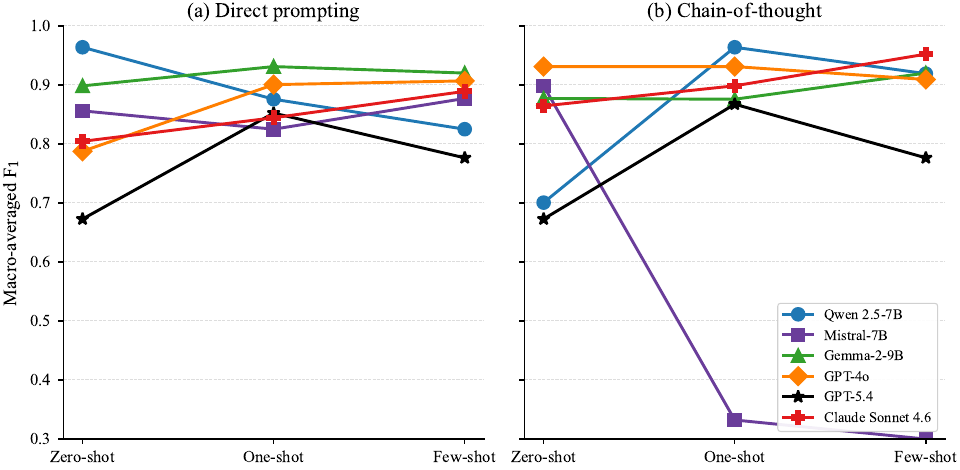}
    \caption{Macro-$F_1$ vs.\ prompting strategy on the binary task.
    Direct (left) and CoT (right). Most LLMs exceed $F_1 = 0.85$ in
    their best configuration; Mistral~+~CoT at higher ICL counts is
    the conspicuous outlier.}
    \label{fig:strategy_binary}
\end{figure}

Figure~\ref{fig:strategy_binary} shows macro-$F_1$ vs.\ strategy on the
binary task. The within-model trends show that most
LLMs benefit modestly from additional ICL exemplars, and CoT yields
mixed effects (Fig.~\ref{fig:cot_binary}). Mistral's CoT collapse is
even more dramatic in the binary framing, dropping the danger-class
$F_1$ from $0.923$ at zero-shot~+~CoT to $0.091$ at few-shot~+~CoT
(Table~\ref{tab:perclass_binary}). Conversely, Qwen at one-shot~+~CoT
matches its zero-shot direct performance ($F_1 = 0.964$), suggesting
the open-source CoT picture is highly model-specific rather than
uniformly beneficial or harmful. 

\begin{table}[ht]
\centering
\caption{Per-class F$_1$ scores on the binary CTAF-KHAF benchmark
(Nominal vs.\ Danger) across all models, prompting strategies, and
reasoning methods. Best per (model, class) row in \textbf{bold}.}
\label{tab:perclass_binary}
\renewcommand{\arraystretch}{1.10}
\small
\begin{tabular}{l l *{6}{c}}
\toprule
 & & \multicolumn{3}{c}{Direct prompting}
 & \multicolumn{3}{c}{Chain-of-thought} \\
\cmidrule(lr){3-5}\cmidrule(lr){6-8}
Model & Class & ZS & OS & FS & ZS & OS & FS \\
\midrule
\rowcolor{gray!20}\multicolumn{8}{l}{\textit{Open-source}} \\
\midrule
Qwen 2.5-7B & Nominal & \textbf{0.951} & 0.845 & 0.795 & 0.560 & \textbf{0.951} & 0.896 \\
            & Danger  & \textbf{0.976} & 0.906 & 0.855 & 0.841 & \textbf{0.976} & 0.942 \\
\cmidrule(lr){1-8}
Mistral-7B  & Nominal & 0.827 & 0.795 & 0.849 & \textbf{0.873} & 0.517 & 0.508 \\
            & Danger  & 0.885 & 0.855 & 0.904 & \textbf{0.923} & 0.147 & 0.091 \\
\cmidrule(lr){1-8}
Gemma-2-9B  & Nominal & 0.873 & \textbf{0.912} & 0.899 & 0.849 & 0.845 & 0.899 \\
            & Danger  & 0.923 & \textbf{0.950} & 0.941 & 0.904 & 0.906 & 0.941 \\
\midrule
\rowcolor{gray!20}\multicolumn{8}{l}{\textit{Closed-source}} \\
\midrule
GPT-4o            & Nominal & 0.681 & 0.862 & 0.879 & \textbf{0.912} & \textbf{0.912} & 0.886 \\
                  & Danger  & 0.894 & 0.938 & 0.934 & \textbf{0.950} & \textbf{0.950} & 0.932 \\
\cmidrule(lr){1-8}
GPT-5.4           & Nominal & 0.488 & 0.784 & 0.667 & 0.488 & \textbf{0.808} & 0.667 \\
                  & Danger  & 0.857 & 0.920 & 0.886 & 0.857 & \textbf{0.926} & 0.886 \\
\cmidrule(lr){1-8}
Claude Sonnet 4.6 & Nominal & 0.733 & 0.778 & 0.847 & 0.814 & 0.857 & \textbf{0.935} \\
                  & Danger  & 0.875 & 0.910 & 0.930 & 0.915 & 0.939 & \textbf{0.968} \\
\bottomrule
\end{tabular}
\end{table}

\begin{table}[ht]
\centering
\caption{Binary classification results (Nominal vs.\ Danger) on the
CTAF-KHAF benchmark. Best per (model, metric) row in \textbf{bold}.
AUROC is computed against the danger class using token logprobs where
available.}
\label{tab:main_binary}
\renewcommand{\arraystretch}{1.10}
\small
\begin{tabular}{l l *{6}{c}}
\toprule
 & & \multicolumn{3}{c}{Direct prompting}
 & \multicolumn{3}{c}{Chain-of-thought} \\
\cmidrule(lr){3-5}\cmidrule(lr){6-8}
Model & Metric & ZS & OS & FS & ZS & OS & FS \\
\midrule
\rowcolor{gray!20}\multicolumn{8}{l}{\textit{Open-source}} \\
\midrule
Qwen 2.5-7B & Macro-F$_1$ & \textbf{0.964} & 0.876 & 0.825 & 0.700 & \textbf{0.964} & 0.919 \\
            & Accuracy    & \textbf{0.968} & 0.883 & 0.830 & 0.766 & \textbf{0.968} & 0.926 \\
            & AUROC       & \textbf{0.995} & 0.973 & 0.983 & 0.994 & 0.979 & 0.984 \\
\cmidrule(lr){1-8}
Mistral-7B  & Macro-F$_1$ & 0.856 & 0.825 & 0.877 & \textbf{0.898} & 0.332 & 0.300 \\
            & Accuracy    & 0.862 & 0.830 & 0.883 & \textbf{0.904} & 0.383 & 0.362 \\
            & AUROC       & 0.897 & 0.968 & \textbf{0.984} & 0.942 & 0.548 & 0.425 \\
\cmidrule(lr){1-8}
Gemma-2-9B  & Macro-F$_1$ & 0.898 & \textbf{0.931} & 0.920 & 0.877 & 0.876 & 0.920 \\
            & Accuracy    & 0.904 & \textbf{0.936} & 0.926 & 0.883 & 0.883 & 0.926 \\
            & AUROC       & 0.951 & 0.979 & \textbf{0.982} & 0.978 & 0.969 & 0.968 \\
\midrule
\rowcolor{gray!20}\multicolumn{8}{l}{\textit{Closed-source}} \\
\midrule
GPT-4o            & Macro-F$_1$ & 0.787 & 0.900 & 0.907 & \textbf{0.931} & \textbf{0.931} & 0.909 \\
                  & Accuracy    & 0.840 & 0.915 & 0.915 & \textbf{0.936} & \textbf{0.936} & 0.915 \\
                  & AUROC       & \textbf{0.991} & 0.983 & 0.976 & 0.977 & 0.988 & 0.978 \\
\cmidrule(lr){1-8}
GPT-5.4           & Macro-F$_1$ & 0.672 & 0.852 & 0.776 & 0.672 & \textbf{0.867} & 0.776 \\
                  & Accuracy    & 0.777 & 0.883 & 0.830 & 0.777 & \textbf{0.894} & 0.830 \\
                  & AUROC       & 0.902 & 0.935 & 0.924 & 0.935 & \textbf{0.952} & 0.915 \\
\cmidrule(lr){1-8}
Claude Sonnet 4.6 & Macro-F$_1$ & 0.804 & 0.844 & 0.889 & 0.864 & 0.898 & \textbf{0.952} \\
                  & Accuracy    & 0.830 & 0.872 & 0.904 & 0.883 & 0.915 & \textbf{0.957} \\
                  & AUROC       & 0.937 & 0.974 & 0.964 & 0.961 & 0.982 & \textbf{0.994} \\
\bottomrule
\end{tabular}
\end{table}

\subsection{Confusion Structure}
Table~\ref{tab:cm_binary} summarizes the $2 \times 2$ confusion
structure for the best-performing configuration of each LLM, broken
out into true-negative (TN), false-positive (FP), false-negative (FN),
and true-positive (TP) counts together with the safety-relevant
\emph{FN rate} (missed-danger) and \emph{FP rate} (false-alarm).
A clear difference emerges between the larger closed-source models
and the smaller open-source models. GPT-5.4 exhibits a conservative
over-alerting behavior, achieving zero missed-danger cases (FN rate
$0.0\%$) at the cost of a high false-alarm rate ($32.3\%$ of nominal
scenarios misclassified as danger). Claude Sonnet~4.6 and GPT-4o
maintain a more balanced trade-off, with low FN rates ($3.2\%$ and
$9.5\%$ respectively) while keeping false alarms relatively limited
($6.5\%$ for Claude and $0.0\%$ for GPT-4o). In contrast, the smaller
open-source models tend to under-alert rather than over-alert.
Mistral-7B misses $14.3\%$ of true-danger scenarios while issuing no
false alarms, and Gemma-2-9B misses $9.5\%$ with the same
zero-false-alarm profile. Qwen-2.5-7B is the strongest open-source
model, reducing the FN rate to only $1.6\%$ while preserving a low
false-positive rate ($6.5\%$). Overall, the table shows that larger
models are generally more willing to issue cautionary danger
predictions, whereas smaller open-source models (with the exception
of Qwen) are more likely to miss hazardous situations.

\begin{table}[ht]
\centering
\caption{Confusion structure for the best run of each LLM on the
binary task (positive class = \textit{danger}). Rates are computed out of
$N=31$ true-nominal and $D=63$ true-danger scenarios in the 94-scenario
test split: TN rate (specificity) $= \mathrm{TN}/N$,
FP rate (false-alarm) $= \mathrm{FP}/N$,
TP rate (recall) $= \mathrm{TP}/D$,
FN rate (missed-danger) $= \mathrm{FN}/D$.
Arrows indicate the desired direction for a safety advisor.}
\label{tab:cm_binary}
\renewcommand{\arraystretch}{1.10}
\small
\begin{tabular}{l l l c c c c}
\toprule
Source & Model & Best run
 & TN rate $\uparrow$ & FP rate $\downarrow$
 & TP rate $\uparrow$ & FN rate $\downarrow$ \\
\midrule
\multirow{3}{*}{\textit{Open-source}}
 & Qwen 2.5-7B       & OS+CoT &  93.5\% &  6.5\% &  98.4\% &  1.6\% \\
 & Mistral-7B        & ZS+CoT & 100.0\% &  0.0\% &  85.7\% & 14.3\% \\
 & Gemma-2-9B        & OS     & 100.0\% &  0.0\% &  90.5\% &  9.5\% \\
\midrule
\multirow{3}{*}{\textit{Closed-source}}
 & GPT-4o            & OS+CoT & 100.0\% &  0.0\% &  90.5\% &  9.5\% \\
 & GPT-5.4           & OS+CoT &  67.7\% & 32.3\% & 100.0\% &  0.0\% \\
 & Claude Sonnet 4.6 & FS+CoT &  93.5\% &  6.5\% &  96.8\% &  3.2\% \\
\bottomrule
\end{tabular}
\end{table}

\subsection{Threshold-Independent Ranking Quality}
\begin{figure}[h!]
    \centering
    \includegraphics[width=\linewidth]{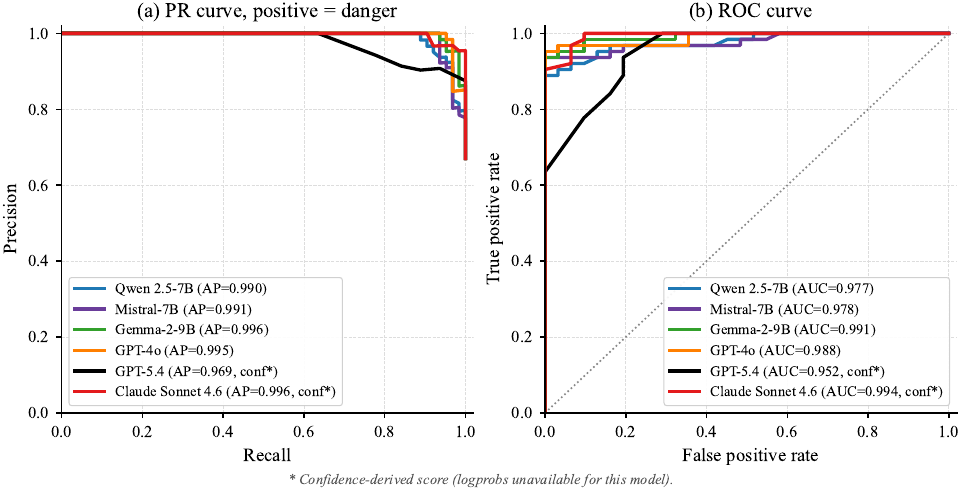}
    \caption{Precision-recall and receiver-operating-characteristic
    curves on the binary task, one curve per LLM (best run by
    macro-$F_1$). $\text{conf}^{*}$ marks score-distribution-based
    fallback for models whose APIs do not expose token logprobs.}
    \label{fig:pr_roc}
\end{figure}

Figure~\ref{fig:pr_roc} reports PR and ROC curves for the best run of
each LLM. All six LLMs hug the top-left corner of the ROC and the
top-right corner of the PR plot, with $\text{AP} \geq 0.97$ and
$\text{AUROC} \geq 0.95$ for every model. Gemma-2-9B and Claude
Sonnet~4.6 attain the highest AP ($0.996$ each); Qwen and GPT-4o are
also at $\geq 0.99$. GPT-5.4 trails at $\text{AP} = 0.969$ /
$\text{AUROC} = 0.952$, in part because its score is necessarily
derived from self-reported confidence rather than token logprobs (the
GPT-5 series rejects the \texttt{logprobs} parameter at the time of
writing); Claude Sonnet~4.6 is similarly flagged. Importantly, the
high AUROC values indicate that the model's score-space rank ordering
of dangerous vs.\ nominal scenarios is reliable across operating
points.

\section{Limitations and Future Work}

A key limitation of the proposed approach lies in the computational
cost and system-level efficiency of LLM/VLMs.
Although we employ a frozen model without fine-tuning, inference
over multimodal inputs remains resource-intensive, posing challenges
for real-time or edge deployment in general aviation settings.
Recent work has shown that efficient deployment of LLMs requires
careful system-level optimization across memory, compute, and
scheduling layers, rather than model design alone
\cite{gogineni2025llms}. Our current architecture does not
incorporate such optimizations, and therefore may not yet meet the
latency and power constraints required for continuous onboard
operation.

A second limitation concerns the data on which the system is
evaluated. Real-world CTAF recordings paired with ground-truth
safety labels are scarce in the public domain, and curating them at
scale would require dedicated audio collection at multiple
non-towered airports together with expert annotation. As a stand-in
we constructed \texttt{CTAF-KHAF-Synthetic}, which captures a
structured 12-category hazard taxonomy at a single airport but
cannot fully reproduce the acoustic noise, accent variation, and
timing irregularities of live radio traffic. The quantitative
results in this article should therefore be read as a measurement on a
controlled, single-airport benchmark rather than a deployment-ready
estimate. A third limitation follows directly from the design of this
synthetic dataset: it does not include per-scenario VFR sectional
imagery or weather radar imagery, so the VLM side of the
architecture cannot be exercised quantitatively on the benchmark.
We were therefore restricted to evaluating LLMs on
the textual subset of the inputs (METAR text and CTAF transcript),
and could only demonstrate the full framework
qualitatively, on a single real recording at KHAF (Sec \ref{sec:qualitative}). A larger benchmark with paired visual modalities would be required for a like-for-like quantitative comparison between LLMs and VLMs in this domain.

Future work includes extending this study into a larger, controlled benchmark in which every scenario is paired with all four data sources (METAR, ADS-B, VFR sectional, and weather radar imagery), at first starting with synthetic data, and then moving into real flight data from the same non-towered airport. This larger more multimodal study would enable a like-for-like quantitative evaluation to our qualitative study. Our use of frozen models was also a limitation, and fine-tuning could improve performance; this remains future work. Future work will also include evaluating the affect of the VFR sectional chart on the models performance, justifying the usage of VLMs over LLMs on the same task.


\section{Conclusion}
This article proposed a VLM-based safety assessment framework for non-towered airports that inputs CTAF radio transcripts, and METAR weather data, ASD-B trajectory data, and VFR sectional images, and produces both a structured safety label and a free-text CTAF-style advisory.  We demonstrated an initial proof of concept with an example at KHAF run with Gemini~2.5~Pro (VLM) which was able to  correctly identify a right-of-way violation that produced a near mid-air collision at KHAF.  We developed a synthetic dataset to quantitatively evaluate performance using the METAR and CTAF data, and benchmarked six frozen LLMs on a binary safety-classification task, where every model exceeded macro $F_1$ of 0.85 in its best configuration and the strongest open-source result tied the strongest closed-source result.

\bibliography{ref} 

\newpage

\section{Appendix}

\begin{figure}[h!]
    \centering
\begin{genprompt}[Dataset Generator Prompts --- \textsc{CTAF-KHAF-Synthetic}]
\footnotesize

\textbf{User message (per scenario).}
For every scenario, the procedurally sampled aircraft, position
events, and METAR are formatted into a single user message:

\begin{tcolorbox}[
    colback=genCode, colframe=genHead!60!black,
    boxrule=0.3pt, arc=2pt,
    left=8pt, right=8pt, top=3pt, bottom=3pt,
    fontupper=\ttfamily\scriptsize]
SCENARIO: \{hazard\_type\} (\{label\}) \\
METAR: \{raw METAR text\} \\
DURATION: \textasciitilde\{duration\}s \\[2pt]
AIRCRAFT: \\
\quad \{callsign A\} (\{type\}) --- radio$|$NORDO \\
\quad \{callsign B\} (\{type\}) --- radio$|$NORDO \\[2pt]
POSITION EVENTS: \\
\quad t=\{t\}s  \{callsign\}  \{phase\}  \{dist\}NM  \{alt\}ft  radio$|$NORDO \\
\quad \ldots \\[2pt]
Write the SRT transcript.
\end{tcolorbox}

This user message is sent to GPT-4o twice, each time paired with one
of the two system prompts shown below.

\smallskip
\textbf{System Prompt 1 --- CTAF Transcript Generation.}
\begin{tcolorbox}[
    colback=genCode, colframe=genHead!60!black,
    boxrule=0.3pt, arc=2pt,
    left=8pt, right=8pt, top=3pt, bottom=3pt,
    fontupper=\ttfamily\scriptsize]
You generate realistic CTAF radio transcripts for Half Moon Bay
Airport (KHAF), runway 30, right-traffic pattern. \\[2pt]
Given exact aircraft positions, write an SRT-format transcript of
pilot radio calls. \\[2pt]
FORMAT (strict): \\
\{index\} \\
\{HH:MM:SS,mmm\} -{}-> \{HH:MM:SS,mmm\} \\
\{text\} \\[2pt]
RULES: \\
- Use NATO phonetic alphabet for letters (Alpha, Bravo\ldots) and ``niner'' for 9. \\
- Each self-announced call: ``Half Moon Bay traffic, [callsign], [position], [runway 30], [intention], Half Moon Bay.'' \\
- NORDO aircraft: only mentioned by other pilots. \\
- Timing: each utterance 3--6 s; gap between calls 3--8 s. Timestamps start at 00:00:00,000. \\
- CRITICAL: total scenario duration MUST be under 90 s; if the last timestamp would exceed 90 s, stop writing calls early. \\
- CRITICAL: write at most 10 lines total; stop at 10 even if not all events are covered. \\
- Return ONLY raw SRT --- no markdown fences, no triple-backtick blocks. \\
- Cover the KEY position events only --- not every single distance update. \\
- Pilots call position at major phase changes: entering downwind, turning base, turning final, short final, going around, clear of runway. \\
- Do NOT have pilots repeat the same position multiple times unless there is a conflict. \\
- For IMC / disoriented pilots: write hesitant, confused speech (``uh'', ``I got\ldots'', corrections). \\
- Return ONLY the SRT content.
\end{tcolorbox}

\smallskip
\textbf{System Prompt 2 --- Ground-Truth Safety Advisory.}
\begin{tcolorbox}[
    colback=genCode, colframe=genHead!60!black,
    boxrule=0.3pt, arc=2pt,
    left=8pt, right=8pt, top=3pt, bottom=3pt,
    fontupper=\ttfamily\scriptsize]
You are an AI aviation safety advisor monitoring CTAF at KHAF
(Half Moon Bay Airport). \\
Write a concise ground-truth safety advisory (2--4 sentences,
\textasciitilde 100--200 words) based on the scenario. \\
Identify aircraft by callsign and type, state their positions
precisely, assess the safety situation, and give recommended actions
if needed. \\
Return ONLY the advisory text.
\end{tcolorbox}
\end{genprompt}

\end{figure}

\subsection{Synthetic Dataset Generation Prompt}
\label{app:gen_prompt}

The \textsc{CTAF-KHAF-Synthetic} benchmark is constructed in two
stages. First, for each of the twelve hazard-type categories in our
taxonomy (e.g., \emph{simultaneous\_final}, \emph{silent\_traffic},
\emph{nominal\_single\_aircraft}), a deterministic Python procedure
samples aircraft callsigns and types, decides which aircraft are
NORDO, generates a sequence of position events at specific
timestamps (entering downwind, turning base, turning final, short
final, going around, clearing the runway, and so on), and draws a
representative METAR for Half Moon Bay Airport (KHAF). Second, these
per-scenario events are packed into a structured user message and
passed to GPT-4o twice, with two different system prompts: once with
the \emph{Transcript} system prompt to produce a strict SRT-format
pilot-radio transcript that follows FAA AC~90-66C phraseology, and
once with the \emph{Advisory} system prompt to produce a
two-to-four-sentence ground-truth safety advisory in FAA-conforming
language. Both LLM calls use temperature $0.7$ and a $1{,}200$-token
output cap, wrapped in retry logic with exponential backoff. The
complete user-message template and the two system prompts are
reproduced in the box above. After generation, every scenario was
reviewed by human experts with general-aviation flight experience to
verify that the radio call sequence is plausible in real CTAF
operations and that the assigned safety label is unambiguous (see
Sec.~\ref{sec:method:dataset} of the main paper). The verified
transcripts are then synthesized into multi-voice MP3 audio using
OpenAI's TTS-1-HD, with a distinct voice assigned to each aircraft
on the frequency so that speaker attribution is preserved in the
audio.

\subsection{Three-Class Classification}
\label{app:3class}

The system prompt used in the three-class classification is shown in
the box below. Aggregated macro-$F_1$ and accuracy values for all six
LLMs, three prompting strategies, and two reasoning protocols are
reported in Table~\ref{tab:main_3class}; per-class $F_1$ for the same
36 conditions is in Table~\ref{tab:perclass_3class}. The three-class task is increased in difficulty: the strongest configuration
(GPT-4o, Few-shot~+~CoT) reaches macro-$F_1 = 0.828$ and accuracy
$0.830$, leaving a clear gap to perfect classification. Closed-source
LLMs dominate the leaderboard---GPT-4o (best configuration $F_1 = 0.828$), Claude Sonnet~4.6 , and GPT-5.4 ($F_1 = 0.820$) all sit in
the $0.82$--$0.83$ band, well above the best open-source results
(Qwen $F_1 = 0.743$, Gemma-2-9B $F_1 = 0.726$, Mistral-7B
$F_1 = 0.640$). The closed-source LLMs also remain consistent across
prompting strategies, whereas the open-source LLMs exhibit substantial
within-model variance.

\begin{figure}[h!]
    \centering

\begin{vlmprompt}[System Prompt --- Three-Class Safety Classification]
You are an automated aviation safety monitoring system for Half Moon Bay
Airport (KHAF), a non-towered airport near San Francisco, California. Your
task is to analyze multimodal flight-operations data and classify the
safety status of the current traffic situation.

\medskip
\textbf{Inputs}
\begin{itemize}[leftmargin=1.4em, topsep=2pt, itemsep=1pt]
    \item METAR weather data for KHAF (raw + decoded)
    \item CTAF radio transcript (SRT format with timestamps)

\end{itemize}

\medskip
\textbf{Task.} Classify the situation as exactly one of \texttt{nominal},
\texttt{warning}, or \texttt{hazard}.

\medskip
\textbf{NOMINAL ---} all is well.
\begin{itemize}[leftmargin=1.4em, topsep=2pt, itemsep=1pt]
    \item All required position calls are present
          (crosswind, downwind, base, final)
    \item Traffic is sequenced and separated, no conflicts
    \item Weather is VMC and appropriate for operations
\end{itemize}

\medskip
\textbf{WARNING ---} a potential problem exists but no collision is
imminent yet.
\begin{itemize}[leftmargin=1.4em, topsep=2pt, itemsep=1pt]
    \item An aircraft flying the wrong pattern direction without conflict
    \item Two aircraft converging on final with separation $>$ 0.5~NM
    \item Missing position calls from one aircraft, no immediate conflict
\end{itemize}
\textit{Key question.} ``Can the pilots resolve this themselves with
standard advisory actions?'' If yes $\Rightarrow$ \textbf{warning}.

\medskip
\textbf{HAZARD ---} a collision or serious incident is imminent or
already occurring.
\begin{itemize}[leftmargin=1.4em, topsep=2pt, itemsep=1pt]
    \item Two aircraft simultaneously on final for the same runway
          ($<$ 0.5~NM)
    \item An aircraft on the runway while another is on short final
    \item Wrong-runway announcement during an active approach
    \item Same altitude and converging --- mid-air collision risk
\end{itemize}
\textit{Key question.} ``Would a CTAF advisory say
\textsc{immediately} or \textsc{safety alert}?''
If yes $\Rightarrow$ \textbf{hazard}.

\medskip
\textbf{Critical distinction.} The difference between
\texttt{warning} and \texttt{hazard} is \textbf{imminence}, not severity.

\medskip
\textbf{Output format.} Respond with \emph{only} the following JSON, no
other text:

\begin{tcolorbox}[
    colback=promptCode, colframe=promptHead!60!black,
    boxrule=0.3pt, arc=2pt,
    left=10pt, right=10pt, top=4pt, bottom=4pt,
    fontupper=\ttfamily\small]
\{\\
\quad "label": "<nominal | warning | hazard>",\\
\quad "confidence": <0.0--1.0>,\\
\quad "reasoning": "<one sentence stating the key safety factor>"\\
\}
\end{tcolorbox}
\end{vlmprompt}
\end{figure}

\begin{table}[ht]
\centering
\caption{Three-class classification results (Nominal / Warning / Hazard)
on the CTAF-KHAF benchmark. Best per (model, metric) row in
\textbf{bold}.}
\label{tab:main_3class}
\renewcommand{\arraystretch}{1.10}
\small
\begin{tabular}{l l *{6}{c}}
\toprule
 & & \multicolumn{3}{c}{Direct prompting}
 & \multicolumn{3}{c}{Chain-of-thought} \\
\cmidrule(lr){3-5}\cmidrule(lr){6-8}
Model & Metric & ZS & OS & FS & ZS & OS & FS \\
\midrule
\rowcolor{gray!20}\multicolumn{8}{l}{\textit{Open-source}} \\
\midrule
Qwen 2.5-7B & Macro-F$_1$ & 0.687 & 0.615 & 0.515 & 0.416 & 0.692 & \textbf{0.743} \\
            & Accuracy    & 0.681 & 0.628 & 0.553 & 0.479 & 0.681 & \textbf{0.755} \\
\cmidrule(lr){1-8}
Mistral-7B  & Macro-F$_1$ & 0.560 & 0.570 & \textbf{0.640} & 0.390 & 0.504 & 0.520 \\
            & Accuracy    & 0.585 & 0.638 & \textbf{0.681} & 0.468 & 0.617 & 0.628 \\
\cmidrule(lr){1-8}
Gemma-2-9B  & Macro-F$_1$ & 0.663 & 0.666 & \textbf{0.726} & 0.456 & 0.616 & 0.687 \\
            & Accuracy    & 0.670 & 0.670 & \textbf{0.755} & 0.500 & 0.606 & 0.681 \\
\midrule
\rowcolor{gray!20}\multicolumn{8}{l}{\textit{Closed-source}} \\
\midrule
GPT-4o            & Macro-F$_1$ & 0.791 & 0.766 & 0.789 & 0.764 & 0.781 & \textbf{0.828} \\
                  & Accuracy    & 0.798 & 0.777 & 0.798 & 0.766 & 0.777 & \textbf{0.830} \\
\cmidrule(lr){1-8}
GPT-5.4           & Macro-F$_1$ & 0.782 & \textbf{0.820} & 0.818 & 0.747 & 0.808 & 0.759 \\
                  & Accuracy    & 0.777 & \textbf{0.819} & \textbf{0.819} & 0.745 & 0.809 & 0.766 \\
\cmidrule(lr){1-8}
Claude Sonnet 4.6 & Macro-F$_1$ & 0.792 & 0.715 & 0.675 & 0.770 & 0.761 & \textbf{0.827} \\
                  & Accuracy    & 0.787 & 0.723 & 0.713 & 0.777 & 0.766 & \textbf{0.830} \\
\bottomrule
\end{tabular}
\end{table}
\begin{table}[ht]
\centering
\caption{Per-class F$_1$ scores on the three-class CTAF-KHAF benchmark
across all models, prompting strategies, and reasoning methods. Best per
(model, class) row in \textbf{bold}.}
\label{tab:perclass_3class}
\renewcommand{\arraystretch}{1.10}
\small
\begin{tabular}{l l *{6}{c}}
\toprule
 & & \multicolumn{3}{c}{Direct prompting}
 & \multicolumn{3}{c}{Chain-of-thought} \\
\cmidrule(lr){3-5}\cmidrule(lr){6-8}
Model & Class & ZS & OS & FS & ZS & OS & FS \\
\midrule
\rowcolor{gray!20}\multicolumn{8}{l}{\textit{Open-source}} \\
\midrule
Qwen 2.5-7B & Nominal & 0.828 & 0.857 & 0.756 & 0.512 & 0.852 & \textbf{0.968} \\
            & Warning & 0.545 & 0.469 & 0.269 & 0.569 & 0.613 & \textbf{0.709} \\
            & Hazard  & \textbf{0.688} & 0.519 & 0.519 & 0.167 & 0.610 & 0.553 \\
\cmidrule(lr){1-8}
Mistral-7B  & Nominal & 0.827 & 0.805 & 0.861 & 0.488 & 0.899 & \textbf{0.921} \\
            & Warning & 0.400 & 0.200 & 0.356 & 0.561 & 0.614 & \textbf{0.638} \\
            & Hazard  & 0.453 & \textbf{0.704} & \textbf{0.704} & 0.121 & 0.000 & 0.000 \\
\cmidrule(lr){1-8}
Gemma-2-9B  & Nominal & \textbf{0.939} & 0.769 & 0.909 & 0.591 & 0.769 & 0.852 \\
            & Warning & 0.508 & 0.475 & 0.489 & \textbf{0.561} & 0.431 & 0.516 \\
            & Hazard  & 0.542 & 0.753 & \textbf{0.779} & 0.216 & 0.648 & 0.694 \\
\midrule
\rowcolor{gray!20}\multicolumn{8}{l}{\textit{Closed-source}} \\
\midrule
GPT-4o            & Nominal & 0.899 & 0.899 & 0.899 & 0.968 & 0.915 & \textbf{0.969} \\
                  & Warning & 0.667 & 0.618 & 0.655 & 0.694 & 0.687 & \textbf{0.733} \\
                  & Hazard  & 0.806 & 0.781 & \textbf{0.812} & 0.630 & 0.742 & 0.781 \\
\cmidrule(lr){1-8}
GPT-5.4           & Nominal & 0.836 & \textbf{0.900} & 0.842 & 0.792 & 0.897 & 0.871 \\
                  & Warning & 0.696 & \textbf{0.730} & \textbf{0.730} & 0.636 & 0.712 & 0.607 \\
                  & Hazard  & 0.812 & 0.831 & \textbf{0.882} & 0.812 & 0.817 & 0.800 \\
\cmidrule(lr){1-8}
Claude Sonnet 4.6 & Nominal & 0.833 & 0.825 & 0.886 & 0.899 & 0.862 & \textbf{0.923} \\
                  & Warning & 0.714 & 0.536 & 0.400 & 0.644 & 0.621 & \textbf{0.733} \\
                  & Hazard  & \textbf{0.828} & 0.783 & 0.740 & 0.767 & 0.800 & 0.825 \\
\bottomrule
\end{tabular}
\end{table}

\begin{figure}[h!]
    \centering
    \includegraphics[width=0.85\linewidth]{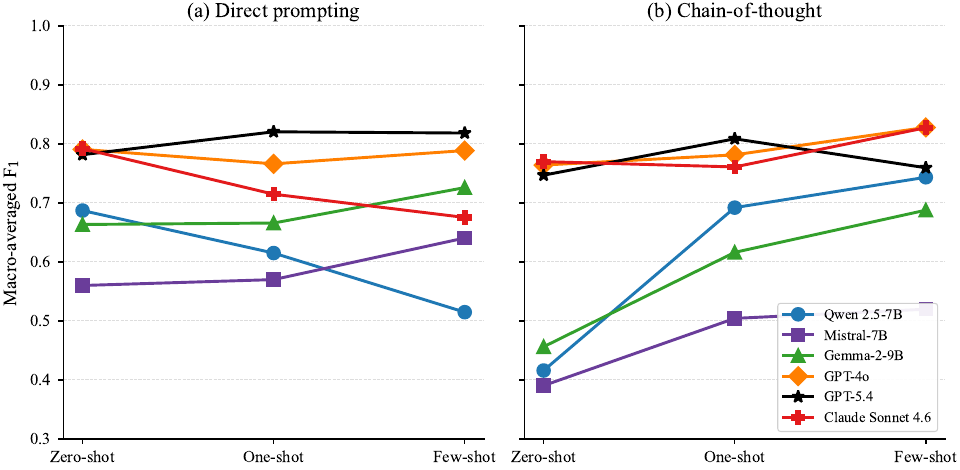}
    \caption{Macro-$F_1$ vs.\ prompting strategy for the three-class
    task. Left panel: direct prompting; right panel: chain-of-thought.
    Closed-source LLMs (GPT-4o, GPT-5.4, Claude) are clustered in the
    top band and respond modestly to ICL; open-source LLMs show higher
    variance and CoT-induced regressions.}
    \label{fig:strategy_3class}
\end{figure}

Figure~\ref{fig:strategy_3class} plots macro-$F_1$ as a function of
prompting strategy, separating direct prompting from chain-of-thought.
Adding ICL exemplars improves most LLMs under direct prompting
(e.g., Gemma-2-9B improves from $0.663$ at zero-shot to $0.726$ at
few-shot), but degrades performance for others (e.g., Claude Sonnet 4.6
drops from $0.792$ at zero-shot to $0.675$ at few-shot).
Under CoT, the trend reverses for some models: Mistral-7B's
hazard-class $F_1$ drops from $0.453$ at zero-shot~+~CoT to
$0.000$ at one-shot~+~CoT and few-shot~+~CoT
(Table~\ref{tab:perclass_3class}), indicating that the additional
exemplars compounded a failure mode rather than mitigating it.

\begin{figure}[h!]
    \centering
    \includegraphics[width=\linewidth]{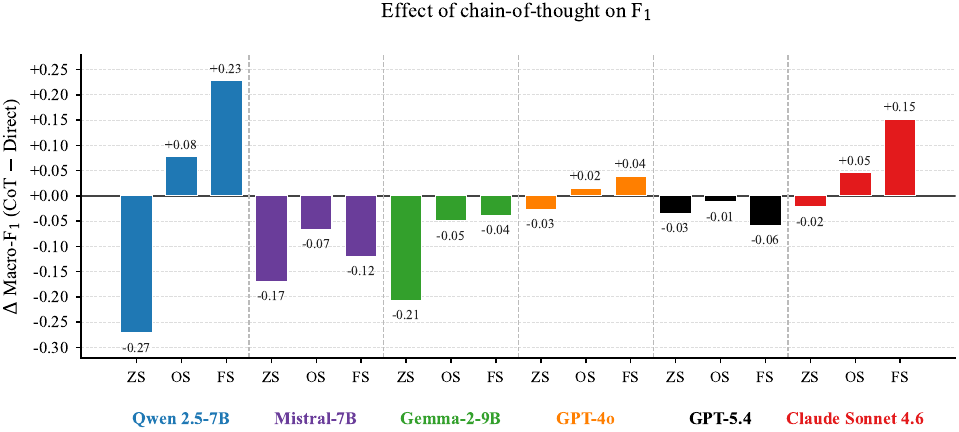}
    \caption{Per-condition macro-$F_1$ change from adding chain-of-thought
    on the three-class task. Qwen benefits at higher ICL counts;
    closed-source LLMs see small mixed effects; Mistral and Gemma both
    regress, with Mistral collapsing entirely on the hazard class at
    the higher-shot CoT settings.}
    \label{fig:cot_3class}
\end{figure}

\begin{figure}[h!]
    \centering
    \includegraphics[width=\linewidth]{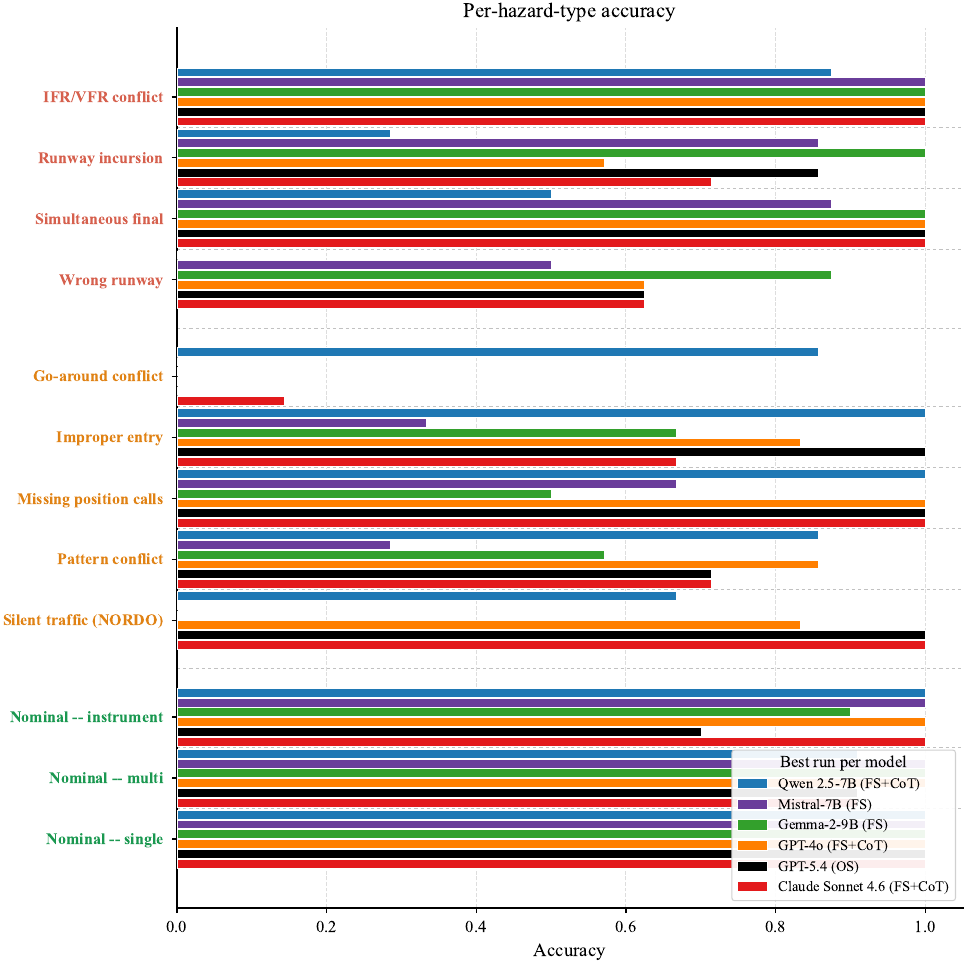}
    \caption{Per-hazard-type accuracy on the three-class task, showing
    the best run per LLM. Hazard categories on the y-axis are colored
    by their ground-truth safety class.}
    \label{fig:perhazard_3class}
\end{figure}

Figure~\ref{fig:cot_3class} reports the per-condition $F_1$ delta
between CoT and direct prompting. The pattern is striking: CoT helps
Qwen at the few-shot setting ($\Delta F_1 = +0.23$) and offers small
gains for closed-source LLMs, but it consistently hurts Mistral
(deltas of $-0.17$ to $-0.12$) and Gemma ($-0.21$ to $-0.04$). On
inspection of individual records, Mistral's CoT failure is not a
parsing or reasoning error: turn-1 reasoning frequently identifies the
correct hazard, but the JSON-extraction turn discards it and emits a
generic ``nominal'' or ``warning'' template. This is a known weakness
of small instruction-tuned LLMs on multi-turn structured-output tasks. 
Figure~\ref{fig:perhazard_3class} shows per-hazard accuracy for the
best run of each LLM. Two patterns emerge: (i)~unambiguous hazards
(\textit{simultaneous final}, \textit{nominal single-aircraft},
\textit{nominal instrument approach}) are solved by every LLM with
accuracy near $1.0$; (ii)~scenarios that hinge on \emph{imminence}
rather than overt symptoms (\textit{runway incursion risk},
\textit{go-around conflict}) yield highly model-dependent accuracy,
ranging from $\sim 0.0$ to $1.0$ across the six LLMs. This is
consistent with our system-prompt definition where the
warning-versus-hazard distinction is explicitly imminence-based and
therefore most exposed to model-side disagreement on subtle scenarios.

\begin{figure}[h!]
    \centering
    \includegraphics[width=\linewidth]{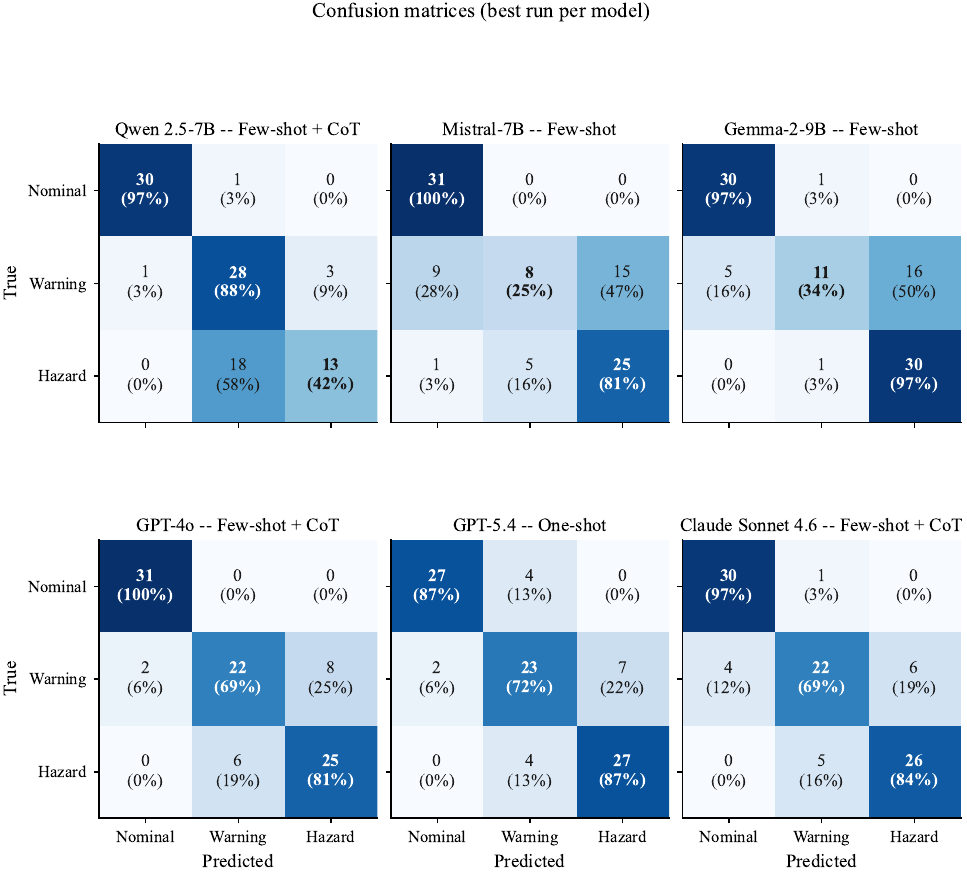}
    \caption{Confusion matrices for the best run of each LLM on the
    three-class task. Diagonal cells are bolded; row-normalized
    percentages indicate per-class recall.}
    \label{fig:cm_3class}
\end{figure}

\begin{figure}[h!]
    \centering
    \includegraphics[width=\linewidth]{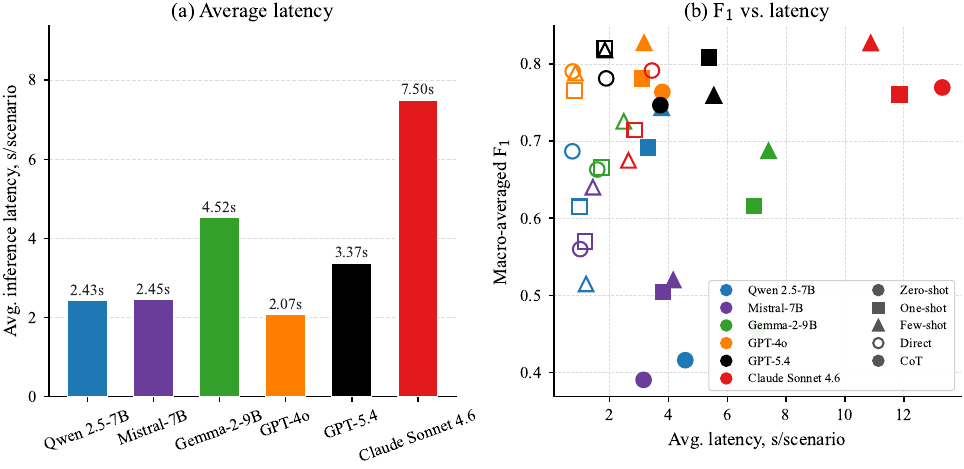}
    \caption{Inference latency on the three-class task. Left: average
    latency per scenario, by LLM. Right: macro-$F_1$ vs.\ latency
    scatter, with marker shape indicating prompting strategy and fill
    indicating CoT vs.\ direct.}
    \label{fig:latency}
\end{figure}

The confusion-matrix grid in Fig.~\ref{fig:cm_3class} shows how each
LLM's errors are distributed. Across the closed-source LLMs (GPT-4o,
GPT-5.4, Claude), the dominant error mode is \emph{warning-as-hazard}
(over-conservative classification), which is operationally desirable
in a safety advisor. Open-source LLMs exhibit more scattered confusion,
with Mistral in particular under-predicting hazard. Figure~\ref{fig:latency} reports average inference latency per
scenario. Latency is measured as the wall-clock time of one full inference call per scenario, averaged across the 94 test scenarios; for chain-of-thought runs it sums both the reasoning turn and the JSON-extraction turn. The within-model spread across prompting strategies in Fig.~\ref{fig:latency}(b) reflects prompt-length scaling: few-shot prompts contain six exemplars, which lengthens the prefill stage and reduces KV-cache reuse across scenarios. GPT-4o is the fastest model overall
($\approx 2.07$~s/scenario), and Qwen and Mistral are the fastest
open-source LLMs ($\approx 2.4$~s/scenario each). Among the
open-source models, Gemma-2-9B is the slowest
($\approx 4.5$~s/scenario), reflecting its larger parameter count
and the lack of cache reuse imposed by the model's chat template.
The slowest model overall is Claude Sonnet~4.6
($\approx 7.5$~s/scenario), well above every other LLM in our
benchmark. The latency vs.\ macro-$F_1$ scatter in the right panel
shows that GPT-4o offers the most favorable cost-quality trade-off
in our deployment regime, since the higher-quality alternatives
(Claude and the few-shot+CoT runs of GPT-5.4) sit further to the
right on the latency axis.

\subsection{Ablation Studies}
\label{app:ablations}

We run three ablations targeting different stages of the pipeline:
ASR quality (Section~\ref{sec:abl:asr}), additive audio noise
(Section~\ref{sec:abl:noise}), and direct masking of the transcript
text (Section~\ref{sec:abl:mask}). All ablations use the three
open-source LLMs (Qwen 2.5-7B, Mistral-7B, Gemma-2-9B) on the same
94-scenario test set, with all main-experiment hyperparameters held
fixed. Closed-source LLMs are excluded from ablations to bound API
cost. As in Sec.~\ref{sec:exp}, only textual inputs (METAR plus CTAF
transcript) are passed to the models, so the LLM terminology is
appropriate throughout.

\subsubsection{ASR Quality}
\label{sec:abl:asr}

\begin{figure}[h!]
    \centering
    \includegraphics[width=0.85\linewidth]{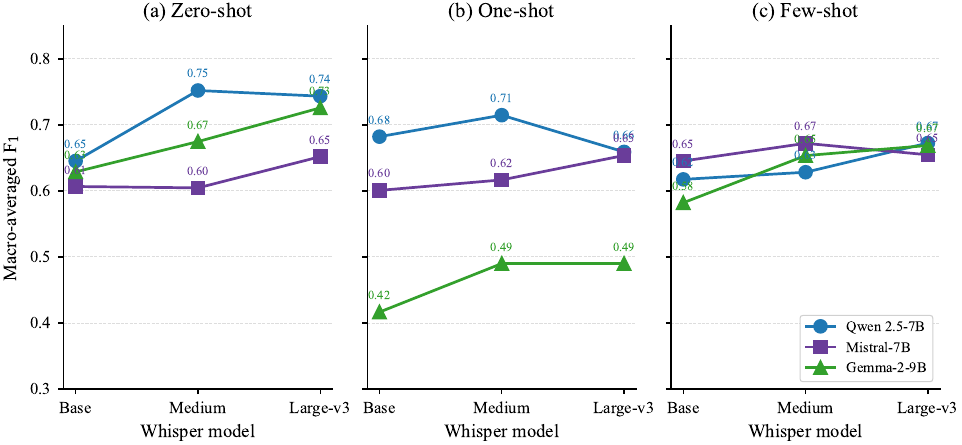}
    \caption{Macro-$F_1$ vs.\ Whisper size for each open-source LLM,
    one panel per prompting strategy.}
    \label{fig:abl_asr_compare}
\end{figure}

We re-transcribe each scenario's clean audio with three Whisper
sizes---\textsf{base} (74M parameters), \textsf{medium} (769M), and
\textsf{large-v3} (1.55B; the default in the main experiments)---and re-evaluate each open-source LLM on each transcript under all three
prompting strategies. Figure~\ref{fig:abl_asr_compare} reports macro-$F_1$ as a function of
Whisper size for each open-source LLM under zero-shot, one-shot, and
few-shot prompting. Under zero-shot prompting, all three LLMs improve as the Whisper model
scales from \textsf{base} to \textsf{medium} and
\textsf{large-v3}, indicating that better transcription quality
consistently benefits downstream hazard classification in the absence
of in-context examples. Qwen-2.5-7B and Gemma-2-9B show the clearest
positive trends, while Mistral-7B exhibits a smaller but still
consistent improvement with increasing ASR size.

Under one-shot prompting, the trends become more model-dependent.
Gemma-2-9B and Mistral-7B continue to improve as the Whisper model
size increases, suggesting that both models are still able to benefit
from incremental ASR gains in the presence of limited prompting
examples. In contrast, Qwen-2.5-7B degrades as the ASR model becomes
larger, indicating that improved transcription quality does not always
translate into improved downstream reasoning once one-shot prompting
is introduced.

Under few-shot prompting, Qwen-2.5-7B and Gemma-2-9B again improve
with increasing Whisper size, although the gains remain relatively
modest. Mistral-7B, however, becomes largely insensitive to ASR scale,
showing only minor fluctuations without a consistent trend: performance
improves slightly at the medium Whisper size before returning close to
its original level with \textsf{large-v3}. This suggests that the
model's downstream behavior is dominated more by prompting dynamics
than by transcription fidelity.

\begin{figure}[h!]
    \centering
    \includegraphics[width=0.65\linewidth]{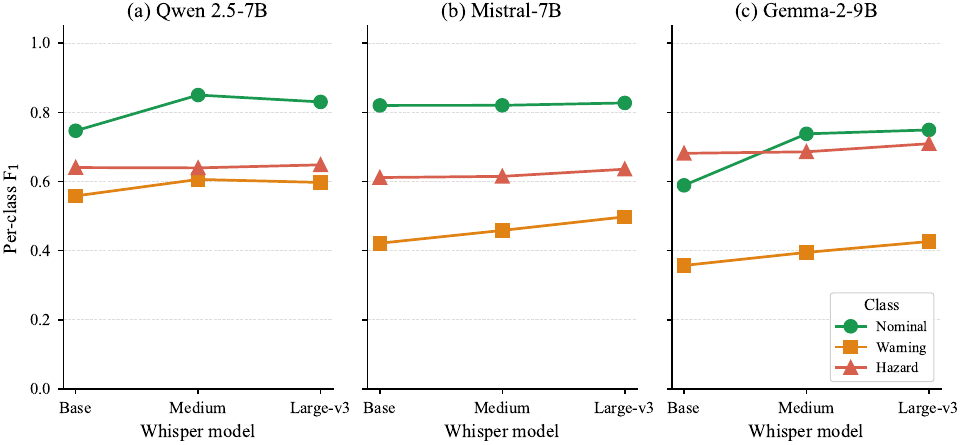}
    \caption{Per-class $F_1$ vs.\ Whisper size, averaged over prompting
    strategy. Color encodes safety class (green = nominal,
    orange = warning, red = hazard).}
    \label{fig:abl_asr_perclass}
\end{figure}

Figure~\ref{fig:abl_asr_perclass} further shows that ASR scaling does
not affect all safety classes equally. Qwen primarily benefits in the
\texttt{nominal} class as Whisper size increases, while the
\texttt{warning} and \texttt{hazard} classes remain comparatively
stable. Mistral exhibits little class-wise variation across all ASR
configurations, reinforcing its overall insensitivity to ASR scale.
Gemma shows the strongest dependence on ASR size, particularly for the
\texttt{nominal} and \texttt{hazard} classes, whereas
\texttt{warning} detection remains consistently more difficult.

Overall, the results suggest that larger ASR models provide only
limited downstream benefits for CTAF hazard classification. Even the
smallest Whisper variant preserves sufficient semantic information for
most traffic situations, and the primary performance bottleneck
appears to lie in the LLM's reasoning and class-boundary calibration
rather than in speech recognition quality itself.

Figure~\ref{fig:abl_asr_perclass} disaggregates per safety class.
Per-class $F_1$ is also nearly invariant under ASR-size choice, with
the warning class showing the most variability---consistent with the
warning class being the bottleneck class in the main experiments
above.

\subsubsection{Audio Noise Robustness}
\label{sec:abl:noise}

\begin{figure}[h!]
    \centering
    \includegraphics[width=0.65\linewidth]{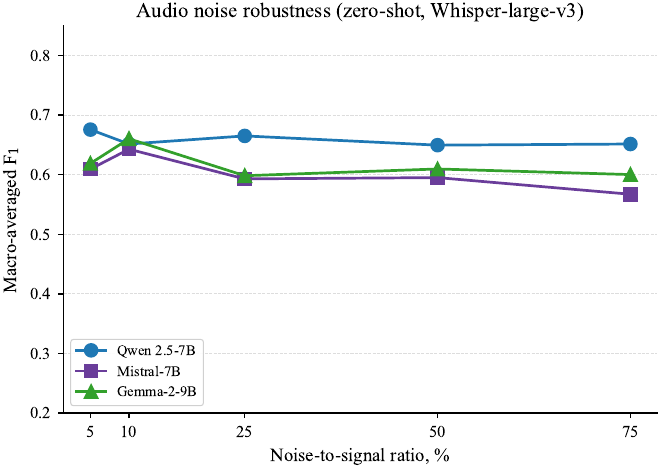}
    \caption{Macro-$F_1$ vs.\ noise-to-signal ratio for each open-source
    LLM under zero-shot prompting and Whisper-Large-v3 transcription.}
    \label{fig:abl_noise_curve}
\end{figure}

We inject additive white Gaussian noise into each scenario's clean
audio at five noise-to-signal ratios (NSR): 5\%, 10\%, 25\%, 50\%, and
75\%. The corrupted audio is transcribed with Whisper-Large-v3 (held fixed) and classified under zero-shot prompting only. Figure~\ref{fig:abl_noise_curve} reports macro-$F_1$ as a function of
noise-to-signal ratio (NSR) for each open-source LLM under zero-shot
prompting with Whisper-\textsf{large-v3} transcription. Overall, all
three LLMs remain relatively stable as audio noise increases,
indicating that the CTAF hazard-classification pipeline is reasonably
robust to moderate transcription degradation caused by noisy radio
communication.

Qwen-2.5-7B exhibits the strongest overall robustness to noise. Its
performance remains largely stable across the full NSR range, with
only minor fluctuations as noise increases. This suggests that Qwen is
able to maintain reliable downstream reasoning even when the ASR input
becomes progressively noisier.

\begin{figure}[h!]
    \centering
    \includegraphics[width=.75\linewidth]{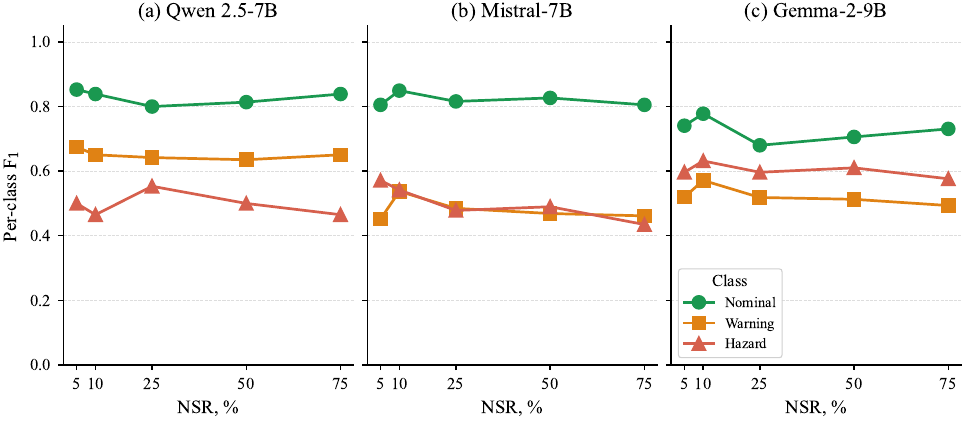}
    \caption{Per-class $F_1$ vs.\ NSR, by LLM. The warning class
    (orange) degrades fastest under high noise.}
    \label{fig:abl_noise_perclass}
\end{figure}

Mistral-7B and Gemma-2-9B show greater sensitivity to increasing
noise. Both models experience an initial improvement at low NSR levels
before degrading as noise increases further. After this early peak,
their performance gradually stabilizes at a lower level across the
higher-noise settings. This behavior suggests that small amounts of
perturbation do not significantly disrupt the semantic structure of
the CTAF transcripts, but heavier degradation eventually reduces the
LLM's ability to consistently separate safety classes. Figure~\ref{fig:abl_noise_perclass} further shows that noise affects
the three safety classes differently. Across all LLMs, the
\texttt{nominal} class remains the most stable under increasing noise,
indicating that routine traffic situations are comparatively easy to
preserve even when transcription quality deteriorates. In contrast,
the \texttt{warning} and \texttt{hazard} classes exhibit greater
variation, particularly for Gemma-2-9B and Mistral-7B. Hazard-related
performance tends to fluctuate more strongly as NSR increases,
suggesting that subtle linguistic cues associated with conflict
imminence are more vulnerable to ASR degradation than standard traffic
phraseology.

Qwen-2.5-7B again demonstrates the most stable class-wise behavior,
with relatively minor changes across all three safety categories.
Mistral-7B exhibits moderate instability in the \texttt{warning} and
\texttt{hazard} classes as noise increases, while Gemma-2-9B shows
the clearest degradation trend under higher NSR conditions,
particularly for hazard detection. Overall, the results indicate that the open-source LLM pipeline is
fairly resilient to noisy CTAF audio, especially for nominal traffic
situations. While increased noise can reduce the reliability of
warning and hazard detection, the degradation remains gradual rather
than catastrophic, suggesting that downstream LLM reasoning retains
substantial robustness even under imperfect ASR conditions.

\subsubsection{Transcript Text Masking}
\label{sec:abl:mask}

\begin{figure}[h!]
    \centering
    \includegraphics[width=0.6\linewidth]{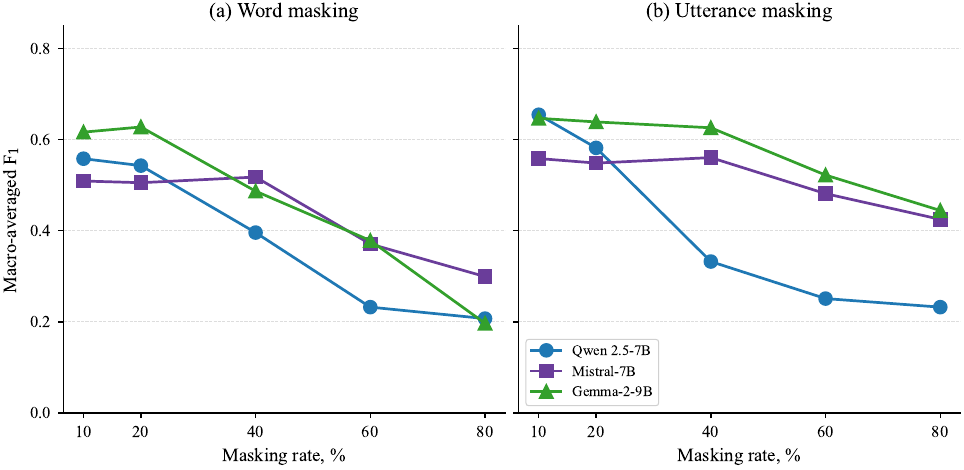}
    \caption{Macro-$F_1$ vs.\ masking rate. Left: word-level masking;
    right: utterance-level masking. Utterance masking is substantially
    more destructive at every rate.}
    \label{fig:abl_mask_curve}
\end{figure}

\begin{figure}[h!]
    \centering
    \includegraphics[width=.95\linewidth]{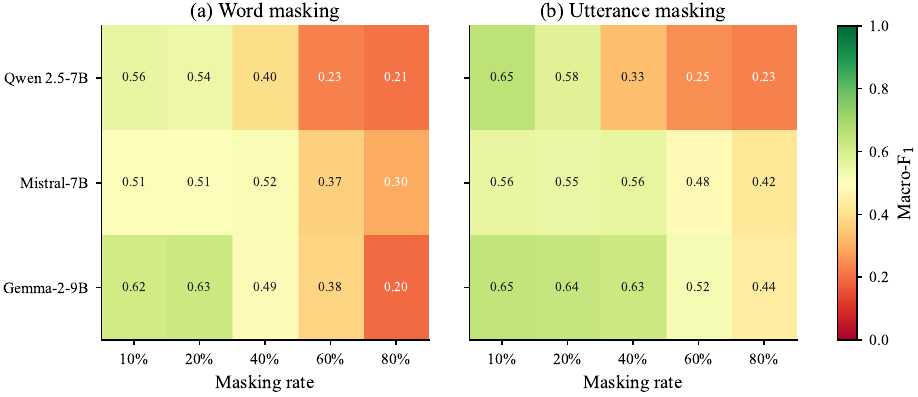}
    \caption{Macro-$F_1$ heatmap across (LLM, mask rate) for word
    (left) and utterance (right) masking. Numerical cell values are
    macro-$F_1$.}
    \label{fig:abl_mask_heatmap}
\end{figure}

To probe tolerance for partial transcript loss directly---without
running the audio chain through Whisper---we mask the ground-truth
transcript at five rates ($r \in \{10, 20, 40, 60, 80\}\%$) under two
schemes. Random fraction of words replaced with a fixed mask token. LLMs still see the conversational structure but with gappy content. And utterance masking, a random fraction of complete radio calls replaced with a placeholder. LLMs see fewer turns overall. Figure~\ref{fig:abl_mask_curve} shows macro-$F_1$ versus mask rate for both masking schemes, while
Fig.~\ref{fig:abl_mask_heatmap} highlights the corresponding
model-by-rate trends. Under word masking, performance generally
degrades gradually as the masking rate increases, indicating that the
LLMs can tolerate the loss of individual words up to moderate masking levels before downstream classification quality deteriorates
substantially. Qwen-2.5-7B and Gemma-2-9B exhibit sharper degradation
at higher masking rates, whereas Mistral-7B remains comparatively more
stable and declines more gradually. 

Utterance masking produces a different but still broadly monotonic
degradation pattern. As the masking rate increases, performance
generally decreases for all three LLMs, although the speed and shape
of the degradation differ from word masking. Because utterance masking
removes complete CTAF transmissions, it disrupts the conversational
and temporal structure used for traffic reasoning rather than only
corrupting local lexical information. Qwen-2.5-7B is particularly
sensitive to this form of masking once moderate amounts of context are
removed, suggesting a strong dependence on sequential conversational
structure. In contrast, Mistral-7B and Gemma-2-9B remain comparatively
more stable under moderate and high utterance masking rates,
indicating that both models can still recover useful traffic-state
information from incomplete conversation histories. Overall, the
results show that utterance masking does not uniformly produce more
severe degradation than word masking; instead, the impact depends on
both the LLM and the type of contextual information being removed.

\begin{figure}[h!]
    \centering
    \includegraphics[width=0.8\linewidth]{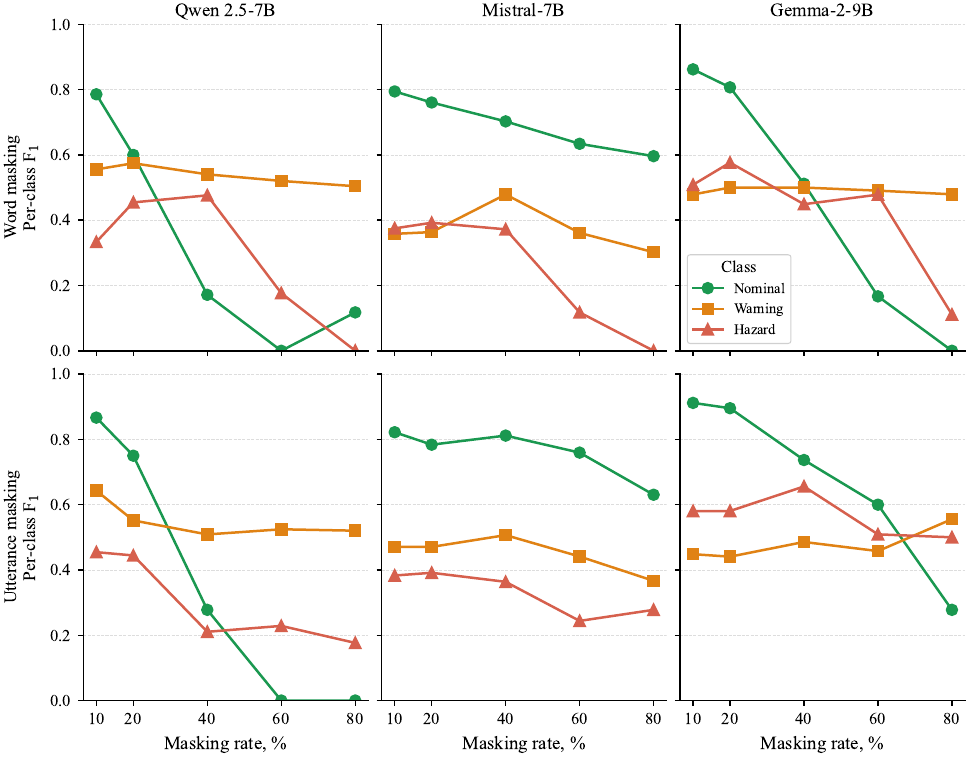}
    \caption{Per-class $F_1$ vs.\ masking rate. Top row: word masking;
    bottom row: utterance masking. Each column is one open-source LLM.}
    \label{fig:abl_mask_perclass}
\end{figure}

Figure~\ref{fig:abl_mask_perclass} further shows that masking affects
the three safety classes differently. Across both masking schemes, the
\texttt{nominal} class experiences the strongest degradation as the
masking rate increases, particularly for Qwen-2.5-7B and Gemma-2-9B.
In contrast, the \texttt{warning} class remains comparatively stable
across most masking levels and models. Hazard detection exhibits
greater variability: Qwen-2.5-7B and Gemma-2-9B show substantial
hazard degradation at high masking rates, while Mistral-7B maintains
more stable hazard performance under utterance masking before
eventually declining at the highest masking levels. These results
suggest that masking primarily disrupts the models' ability to recover
the global conversational state associated with nominal traffic flow,
while hazard-related cues remain partially recoverable even under
incomplete transcript conditions.

\end{document}